\documentclass[5p,twocolumn]{elsarticle}

\usepackage{amsmath,amssymb}
\usepackage{algorithm}
\usepackage{algpseudocode}
\usepackage{graphicx}
\usepackage{float}
\usepackage{array}
\usepackage{tikz}
\usepackage{enumitem}
\usepackage{booktabs}
\usepackage{multirow}
\usepackage{adjustbox}
\usepackage{caption}
\usepackage{siunitx}
\usepackage{threeparttable}
\usepackage{tabularx}
\usepackage{array}
\usetikzlibrary{positioning, arrows.meta, shapes, backgrounds}

\usepackage[hidelinks]{hyperref}

\journal{Knowledge-Based Systems}

\begin{document}

\begin{frontmatter}




\title{MUTEX: A Framework for Toxic Span Detection in Urdu Using URTOX}

\author[pieas]{Inayat Arshad}

\author[pieas]{Fajar Saleem}

\author[pieas]{Ijaz Hussain\corref{cor1}}
\ead{ijazhussain@pieas.edu.pk}

\affiliation[pieas]{
    organization={Department of Computer and Information Sciences, Pakistan Institute of Engineering and Applied Sciences (PIEAS)},
    addressline={Lehtrar Road, Nilore}, 
    city={Islamabad},
    country={Pakistan}
}

\begin{abstract}
Online toxic language detection for Urdu’s 170+ million speakers remains limited because most existing systems rely on sentence-level classification and fail to identify the specific toxic spans within text. This challenge is further amplified by the lack of token-level annotated resources, along with the linguistic complexity of Urdu, including frequent code-switching, informal expressions, and rich morphological variation. These problems are solved in the present study with the presentation of “ MUTEX: the first explainable toxic span detection framework for Urdu”  and “URTOX: the first manually annotated token-level toxic span dataset for Urdu” with 14,342 samples annotated with BIO tagging to improve interpretability and facilitate manual moderation. MUTEX is a system that uses XLM RoBERTa with a CRF layer to perform sequence labeling and the model is tested on multi-domain data of social media, news, and YouTube by token-level F1 to evaluate fine-grained span detectors. The results indicate that MUTEX attains 60\% token-level F1 score, which is the first supervised baseline for Urdu toxic span detection. Further examination has shown that transformer-based models can be effective at implicitly capturing and context-specifically capturing toxicity and are able to address the issues of code-switching and morphological variation. Overall, the work provides the first comprehensive structure for explainable Urdu toxic span detection, including a benchmark dataset, methodological perspective, and practical guidance for constructing more interpretable and accurate content moderation systems in low-resource languages.

\end{abstract}




\end{frontmatter}


\section{Introduction}

The rapid growth of online communication has fundamentally transformed the way people express opinions, emotions, and identities on different digital platforms~\cite{wulczyn2017ex}. The use of social networks such as X (previously Twitter), Instagram, Reddit, Telegram, and YouTube has facilitated information exchange in real-time and has made the world a global village~\cite{davidson2017automated}. Although these channels make communication faster and provide freedom of expression. However, they represent a double-edged sword, as they do not moderate and use any editorial check on the use of toxic content, such as abusive, hateful, and toxic language~\cite{schwartz2020green}. This type of material endangers social division among users, attacks marginalized groups of people, and in severe cases leads to physical violence~\cite{hussain2025survey}. Generally, toxic and hate speech refers to any type of communication—written, spoken, or behavioral—that attacks, threatens, discriminates against, or demeans people or groups of people based on their characteristics, such as religion, ethnicity, race, gender, nationality, or other characteristics of identity~\cite{hussain2025must}. However, even though much has been said, there is no universally adopted single agreed definition of hate speech that has been universally adopted, and the concept is very contextual and culturally dependent ~\cite{schwartz2020green}. This is a major problem for automated detection, especially in multilingual and culturally diverse environments.

Online toxicity is a social issue in polarized societies such as  Pakistan. Users post emotionally colored sentiments of political, religious, and cultural nature on social networks often against minorities and disadvantaged groups~\cite{hussain2025dambert}. Pakistan's national language, Urdu, is written in Nastaliq script and is ranked as one of the top ten most commonly spoken languages in the world ~\cite{hussain2025survey}. Even with its richness in literary tradition, Urdu is grossly under-resourced in terms of natural language processing (NLP), particularly in solving safety-related tasks like toxic and hate speech detection~\cite{schwartz2020green, khan2019urdu}.
Urdu presents unique challenges because of its rich morphology, complex script, context sensitivity, non-standardization in informal writing and regional variation ~\cite{hussain2025dambert}. On social media, these issues are exacerbated by the high rates of code-switching to English, Romanized forms, slang, transliterations, and non-standard spelling~\cite{bhat2017joining}, which ensures that the direct transfer of models centered on the English language cannot be reliable~\cite{khan2019urdu}.

Consequently, current toxicity detection models that have been developed based on high-resource languages are usually not effective when generalized to Urdu and other low-resource languages such as Urdu, Hindi, and Bengali. Most of the current toxicity detection methods do sentence-level classification, which is a classification of an entire text as being either toxic or non-toxic. Although useful for coarse-grained moderation, this type of prediction fails to determine which particular words or phrases cause toxicity. This weakness diminishes interpretability and limits downstream use, such as selective masking, explainable moderation, and accessibility tools~\cite{ref68}. Toxic span detection removes this limitation by determining the precise text segments that are toxic.
In spite of the fact that toxic span detection has been widely researched on English using benchmarks including SemEval-2021 Task~5~\cite{pavlopoulos2021semeval} and later methods of sequence labeling and span modeling, Urdu is largely neglected because of the lack of annotated datasets, baselines models, and evaluation criteria. Subsequent approaches to the detection of English toxic span have reported character-level F1 scores of up to 71.1\%, with recent state-of-the-art models typically achieving the same results~\cite{pavlopoulos2021semeval, ranasinghe2023multilingual,tran2021uit, ref53}; however, meaningful cross-language comparison is difficult due to differences in scripts, annotation schemes, and linguistic complexity.

To address these gaps, we introduce ``MUTEX: the first explainable toxic span detection framework for Urdu'' that performs fine-grained token-level toxicity identification using sequence labeling, enabling precise localization of harmful language while preserving contextual meaning. As part of this work, we also present ``URTOX: a novel manually annotated toxic span dataset consisting of 14,342 Urdu samples'' collected from various social media platforms, i.e. X, Instagram, Reddit, Telegram, Urdu news outlets, and YouTube. Manual annotation with token-level beginning-inside-outside (BIO) tagging was necessary to accurately capture Urdu’s morphological variation, informal expressions, and slang-based toxicity ~\cite{cohen1960coefficient, akhter2024automatic}.

Our framework is based on a hybrid XLM-R+CRF architecture, combining multilingual contextual embeddings~\cite{conneau2020unsupervised, ahmed2025multiclass} with conditional random fields to enforce valid span-level label sequences, a design widely adopted in sequence labeling tasks such as named entity recognition~\cite{lample2016neural, ma2016end}. MUTEX achieves a token-level F1 score of 60\%, which is the first supervised benchmark for Urdu toxic span detection.

Besides performance, MUTEX focuses on transparency by using explainable AI, where token attribution approaches based on gradients are used to highlight the reasons behind words being toxic. Trust and accountability of moderation systems can be enhanced by such explainability, and aligned with recent directions in interpretable NLP.

The main contributions of this work are summarized as follows:
\begin{enumerate}
    \item We curate ``URTOX'', the first manually annotated Urdu toxic span detection dataset, consisting of 14,342 samples collected from social media platforms, Urdu news outlets, and YouTube. The dataset is annotated using token-level BIO tagging to enable fine-grained span supervision. Manual annotation was essential to capture Urdu’s rich morphology, informal expressions, slang, and code-mixed usage. High inter-annotator agreement was achieved, ensuring annotation reliability as discussed in Section~\ref{sec:annotation}.
    \item We propose ``MUTEX'', the first explainable framework for fine-grained toxic span detection in Urdu. MUTEX integrates data preprocessing, sequence labeling, multimodal processing, and explainability into a unified pipeline. The overall system architecture of MUTEX is illustrated in Figure~\ref{fig:system_architecture}.
    \item We set the first supervised benchmark of Urdu toxic span detection to compare several models, such as BiLSTM-CRF, mBERT, XLM-RoBERTa, and XLM-R+CRF. Our top stage model, which has been developed on XLM-R+CRF architecture, attains a token-level F1 score of 60\%, which indicates the efficiency of transformer-based sequence labeling of Urdu. Detailed experimental results are reported in Table~\ref{tab:experimental-results}.
    \item We use explainable AI methods based on gradient-based token attribution, which allows visualizing the decisions of the model in terms of words. This will enable moderators and analysts to know why certain tokens are considered toxic to enhance transparency, trust, and accountability. The qualitative and quantitative explainability analyses are provided in Section~\ref{sec:xai}.
    \item We conduct extensive ablation and transfer learning studies to analyze the impact of architectural choices and preprocessing strategies. Results show that CRF layers contribute a 1--2\% performance gain, while preprocessing yields a cumulative improvement of 8.2\% (Section~\ref{sec:ablation}).

\end{enumerate}

\section{Related Work}

Span-level representations are not optimally captured using the traditional BERT-style models that utilize masked language modeling (MLM) with single-token masking. SpanBERT, proposed by Joshi et al.~\cite{joshi2020spanbert} to mask contiguous random spans and train a predictor of the entire masked span using only the boundary tokens is found to work well with English.

Previously, early toxicity detection systems concentrated on sentence-level or document-level classification and provided one toxicity label to a complete text~\cite{wulczyn2017ex,davidson2017automated}. Although useful for coarse moderation, these methods are not interpretable and do not identify the specific segments that cause the toxic behavior.

Toxic span detection overcomes this limitation by identifying specific character- or token-level spans that are the sources of toxicity—i.e., the exact segments that contribute to the overall toxic nature of the text~\cite{pavlopoulos2021semeval,mathew2021hatexplain}. It is a fine-grained formulation that enhances explainability as well as allows targeted moderation. This task is closely connected with Named Entity Recognition (NER) and was formally introduced in competitions like SemEval-2021 Task 5, which established span-level evaluation measures.

Most models for toxic span detection formulate the task as a sequence tagging problem with BIO tagging schemes~\cite{lample2016neural,ma2016end}. The TRuST model showed that it is possible to use NER-based models to successfully identify toxic spans using capitalisation on contextual embeddings~\cite{ranasinghe2023multilingual}. These approaches are token-level probabilistic maximization using negative log-likelihood:

\begin{equation}
\mathcal{L}_{span} = - \sum_{t=1}^{T} \log P(y_t^{gold} \mid \mathbf{x})
\end{equation}

where $y_t^{gold} \in \{B, I, O\}$ denotes the label assigned to the token at position $t$, and $\mathbf{x}$ is the input Urdu sequence. This formulation allows models to capture token-level dependencies critical for span localization.

Transformer or BiLSTM encoders are often combined with Conditional Random Fields in order to enhance consistency of sequences~\cite{shahzad2025enhancing,ref68}. CRFs are used to model the relationships between neighboring labels, which produce sound span boundaries. The conditional probability of a label sequence in a CRF is defined as:

\begin{equation}
\small P(\mathbf{y} \mid \mathbf{x}) =
\frac{
\exp \Big( \sum_{t=1}^{T} (\psi_{emit}(y_t, \mathbf{h}_t) + \psi_{trans}(y_{t-1}, y_t)) \Big)
}{
\sum_{\mathbf{y'} \in \mathcal{Y}} \exp \Big( \sum_{t=1}^{T} (\psi_{emit}(y'_t, \mathbf{h}_t) + \psi_{trans}(y'_{t-1}, y'_t)) \Big)
}
\end{equation}

where $\psi_{emit}$ is the emission score from hidden XLM-RoBERTa representations and $\psi_{trans}$ is the transition score that imposes valid BIO spans. Recent works indicate that these hybrid and ensemble architectures enhance span detection performance compared to standalone models~\cite{usmani2024roman,saeed2025purutt}.

Transformer architectures such as BERT~\cite{devlin2019bert}, RoBERTa~\cite{ahmed2025multiclass}, and XLM-RoBERTa~\cite{conneau2020unsupervised} have been demonstrated to be state-of-the-art for toxic span detection because they are capable of capturing long-range contextual dependencies. SemEval-2021 demonstrated the effectiveness of transformer models for toxic span detection and proved the superiority of transformers for detecting implicit and context-dependent toxicity~\cite{pavlopoulos2021semeval,kapil2020racism}.

In order to get a clearer insight into the transformer success in detecting toxic spans, Clark et al.~\cite{clark2019does} examined attention patterns of BERT and found that different attention heads are dedicated to detecting different linguistic phenomena, including syntactic dependencies and semantic relationships that play a key role in identifying toxic content. Furthermore, SpanBERT~\cite{joshi2020spanbert} proposed span-based pretraining objectives which directly optimize continuous text spans, showing particular promise for tasks that involve accurate span boundary detection.

Alternative formulations reframe toxic span detection as a question-answering task, such as asking ``Which part of the text is toxic?''~\cite{latif2021deepfake,kugathasan-sumathipala-2021-neural}. The UIT-E10 dot3 model integrated NER and QA approaches and was found to perform highly on the determination of multiple spans in one input~\cite{tran2021uit}. Cross-domain experimentation showed that models trained on a specific platform such as Twitter, are likely to decline their performance by 10--15\% when tested on others such as Reddit or YouTube because of language style and toxicity pattern differences~\cite{caselli2021hatebert,ousidhoum2021multilingual}.

There is limited research on toxic span detection for low-resource cursive languages~\cite{abdel-salam-2022-dialect,kapil2020racism}. The PURUTT corpus proposed a two-step system of Urdu toxicity detection, initially identifying candidate stretches followed by classification, which was effective on compound and idiomatic insults typical of the Urdu language~\cite{shahzad2025survey}. Additional research on cursive languages like Urdu and Arabic showed that NER-based span detection consistently performs better than sentence-level classification, especially with the use of transformer-based contextual representations~\cite{usmani2024roman}.

Although there have been significant developments in toxic language detection, Urdu toxic span detection remains severely limited by critical challenges, including sentence-level classification limitations, single-script corpora bias, limited cross-domain coverage, binary classification approaches, lack of explainability, code-switching complexities affecting approximately 35--40\% of posts, morphological challenges, and limited annotated resources. Our work fills three gaps that are critical, as indicated in the literature review.

First, the current body of research on toxic span detection has been largely constrained to English and high-resource languages, with challenges such as SemEval-2021 Task 5~\cite{pavlopoulos2021semeval},  which has 65–70\% F1 at the benchmark. Although multilingual models, such as XLM-RoBERTa~\cite{conneau2020unsupervised} have demonstrated potential to be used in cross-lingual applications, Urdu is grossly underrepresented in pretraining data, at less than 1\%,and there is no previous research to set span-level baselines in Urdu. Majority of the current literature is based on sentence-level classification and not fine-grained span detection~\cite{wulczyn2017ex,davidson2017automated,latif2021deepfake}, and it does not identify specific toxic words or phrases, which limits interpretability and limits the use of such methods as selective masking and explainable moderation, among others~\cite{mathew2021hatexplain,ref68}. Early efforts to apply toxicity detection to Urdu were done by PURUTT corpus~\cite{shahzad2025survey} which operates at the sentence level, with no ability to identify the specific toxic words and phrases, only that it has toxicity —a limitation our work directly addresses through token-level BIO tagging with high inter-annotator agreement (Cohen's $\kappa = 0.82$, Krippendorff's $\alpha = 0.81$)~\cite{cohen1960coefficient,akhter2024automatic}.

Second, cross-domain performance degradation remains a persistent challenge. It has been demonstrated in the literature that 10–15\% F1 decreases when models trained on one platform are tested on the other platform ~\cite{caselli2021hatebert,ousidhoum2021multilingual}. decreases when models trained on one platform are tested on the other platform 18\% of online Urdu content is in Roman script which sets up a dual-script reality that is not reflected by current datasets~\cite{khan2019urdu, bhat2017joining, malik2010perso, bali2014analysis}. Minimal cross-domain coverage is limited to tweets~\cite{ranasinghe2023multilingual,kugathasan-sumathipala-2021-neural,tran2021uit}, which exposes them to bad generalization to longer and more context-sensitive media such as YouTube comments or news articles~\cite{caselli2021hatebert,ousidhoum2021multilingual,eisenstein2013bad}. We did this systematically in multi-domain training using 3 different sources of Urdu texts, such as social media, news, and YouTube, showing that multi-domain training is outperformed by single-domain training by 1.4\% F1 on average. We also measure domain-specific biases such as script variation, where Roman Urdu led to a 3.2\% F1 decrease, and code-switching effects led to a 1.6\% drop in F1~\cite{ref53,bali2014analysis,solorio2014overview}, and formality gaps, where a difference of 6.1\% is found between social media and news~\cite{eisenstein2013bad}, which offers practical knowledge on the cross-domain toxic span detection.
Third, although hybrid systems with transformers and CRFs have demonstrated 3–5\% better sequence tagging results than plain CRFs with morphologically rich languages with extensive code-switching. behavior~\cite{shahzad2025enhancing,usmani2024roman,ref71}, these methods have not been systematically tested on low-resource toxic span detection. The majority of works use binary classification and do not differentiate between various types of toxicity like hate speech, insults, and profanity~\cite{kapil2020racism,kugathasan-sumathipala-2021-neural, schwartz2020green}, and the morphological complexity of Urdu poses a problem to the detection of span boundaries because toxic words are represented by different morphological forms of the same word conceptualization~\cite{bhat2017joining, sang2003introduction, majumder2020investigating, butt1995urdu, kiros2015skip}. We implement the XLM-RoBERTa+CRF to Urdu toxic span detection the first time and obtain 60.0\% of token-level F1 in the experiment and prove that this time-tested hybrid model is especially efficient when it comes to the linguistic complexity of Urdu. Our ablation experiments estimate the contributions of preprocessing, CRF layers and multi-domain training by 6.2\%, 1.3\%, and 1.4\% respectively, to give empirically informed future directions in low resource language studies.
Lastly, although past research on explainability has concentrated on attention visualization~\cite{clark2019does} or weakly-supervised rationale extraction~\cite{lei2016rationalizing}, models in practice act as black boxes without explainability in many cases~\cite{latif2021deepfake,sundararajan2017axiomatic}, which compromises the trust in the automated systems. We combine gradient-based attribution with integrated gradients~\cite{sundararajan2017axiomatic} to give mathematically rigorous token-level explanations~\cite{ref82, ref69, sundararajan2017axiomatic} which is essential to increase transparency and trust in automated content moderation systems, as this has been defined as a critical requirement by the practitioners applying toxicity detectors in practice.
In summary, our work establishes the first comprehensive framework for Urdu toxic span detection with token-level annotations, systematic cross-domain analysis, and explainable predictions. We propose MUTEX: the first explainable toxic span detection framework for Urdu, addressing rich morphology, dual-script usage, code-switching, and informal expressions through fine-grained token-level identification with BIO tagging. Our multi-domain dataset URTOX (detailed in Section~\ref{sec:dataset}) comprises 14,342 manually annotated samples from Social Media, News, and YouTube, capturing morphological variation, informal expressions, slang, and code-switching that automated tools cannot handle~\cite{cohen1960coefficient,akhter2024automatic,bhat2017joining}. This provides both benchmark results and methodological insights applicable to other underrepresented languages.

\begin{table*}[t]
\centering
\caption{Comparative Analysis of Toxic Span Detection Approaches}
\label{tab:related_works}
\small
\begin{tabularx}{\textwidth}{@{}l l l l X X@{}}
\toprule
\textbf{Ref.} & \textbf{Approach} & \textbf{Model/Method} & \textbf{Lang.} & \textbf{Strengths} & \textbf{Limitations} \\
\midrule

\cite{joshi2020spanbert} & Span Masking Pretraining & SpanBERT & English & Better span boundaries via span-MLM & English only; high pretraining data demand \\
\midrule

\cite{wulczyn2017ex} & Document-level Classification & Logist. Reg. + CNN & English & Fast inference; effective for coarse-grained filtering & Lacks interpretability; cannot identify specific toxic segments \\
\midrule

\cite{davidson2017automated} & Binary Toxicity Classification & Logist. Reg. (n-grams) & English & Simple baseline; interpretable features & Cannot distinguish hate speech from offensive language \\
\midrule

\cite{pavlopoulos2021semeval} & Character-level Span Detection & Transformers & English & Fine-grained detection; benchmark F1: 68.5--70.83\% & Character-level evaluation is computationally expensive \\
\midrule

\cite{mathew2021hatexplain} & Explainable Hate Speech & BERT + Attention & English & Provides rationales; human-annotated explanations & Requires expensive human annotations; sentence-level only \\
\midrule

\cite{lample2016neural} & Sequence Tagging & BiLSTM-CRF & Multi. & Effective for NER; models label dependencies & Requires extensive feature engineering; slower than transformers \\
\midrule

\cite{ma2016end} & End-to-End Seq Labeling & BiLSTM-CNN-CRF & Multi. & No feature engineering; captures char/word features & Computationally expensive; outperformed by transformers \\
\midrule

\cite{ranasinghe2023multilingual} & Multilingual Offensive Det. & XLM-RoBERTa & Multi. & Cross-lingual transfer; low-resource language support & Focus on sentence classification, not span detection \\
\midrule

\cite{ref71} & Sequence Labeling & CRF & -- & Models label dependencies; globally optimal predictions & Requires manual feature engineering; linear dependencies only \\
\midrule

\cite{shahzad2025enhancing} & Neural Sequence Tagging & BiLSTM-CRF & Multi. & Combines neural representations with CRF & Slower inference than standalone models; sequential processing \\
\midrule

\cite{usmani2024roman} & Portuguese NER & BERT-CRF & Portu. & Transfer learning from multilingual BERT; +3\% over BERT & Limited to Portuguese; requires CRF implementation \\
\midrule

\cite{saeed2025purutt} & Multi-task Emotion Det. & Multi-task Ensemble & English & Jointly learns multiple tasks; improved generalization & Complex training; requires multiple labeled datasets \\
\midrule

\cite{devlin2019bert} & Masked Language Modeling & BERT & English & Bidirectional context; transfer learning foundation & Large model size; requires extensive pretraining \\
\midrule

\cite{ahmed2025multiclass} & Optimized BERT Pretrain & RoBERTa & English & Improved over BERT; dynamic masking; larger batches & English-centric; computationally expensive \\
\midrule

\cite{conneau2020unsupervised} & Cross-lingual Pretraining & XLM-RoBERTa & Multi. & Strong multilingual representations; zero-shot transfer & Urdu is severely underrepresented in pretraining ($<$1\% data) \\
\midrule

\cite{kapil2020racism} & Toxic Span w/ Disagreement & Ensemble Transf. & English & Models annotator disagreement; robust predictions & Requires multiple annotations; computational overhead \\
\midrule

\cite{clark2019does} & BERT Attention Analysis & BERT Interpretability & English & Reveals specialized attention heads & Analysis only; does not improve performance \\
\midrule

\cite{latif2021deepfake} & Data Augmentation & Augmented BERT & English & Improves robustness; addresses data scarcity & Augmentation quality varies; may introduce noise \\

\bottomrule
\end{tabularx}
\end{table*}

\section{Methodology}
\label{sec:method}
\subsection{Problem Formulation}

We formulate toxic span detection in Urdu as a supervised sequence labeling problem at the token-level. Given an input text sequence
\[
X = \{x_1, x_2, \dots, x_T\},
\]
where $x_i$ represents the $i$-th token in an Urdu post and $T$ denotes the sequence length, the objective is to identify all contiguous spans of tokens that express toxicity.

Each token $x_i$ is assigned a label $y_i \in \mathcal{Y}$, where the label set $\mathcal{Y} = \{\text{B-TOXIC}, \text{I-TOXIC}, \text{O}\}$ follows the BIO tagging scheme. From the predicted BIO labels, toxic spans are constructed as maximal contiguous subsequences of tokens labeled as \text{B-TOXIC} or \text{I-TOXIC}. For fine-grained evaluation, these token-level spans are further mapped to character-level offsets to obtain a set of predicted toxic character spans $S_A^t$ for each post $t$.

The primary evaluation objective is to maximize span-level F1-score computed over character offsets, which rewards partial overlap between predicted and gold toxic spans and penalizes boundary errors, making it suitable for sparse and imbalanced toxic span detection.

\begin{figure*}[!t]
    \centering
    \includegraphics[width=\textwidth]{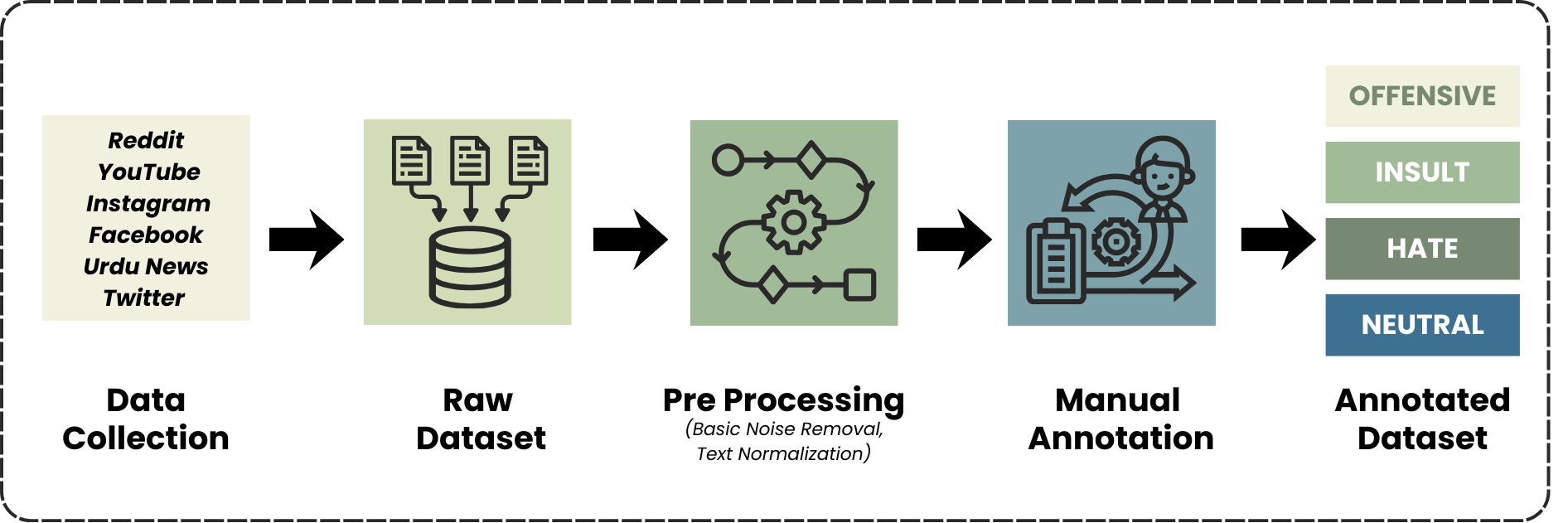}
    \caption{Data collection and annotation pipeline for the Urdu toxic span detection dataset.}
    \label{fig:data_pipeline}
\end{figure*}

\subsection{Data Collection}

To promote generalization across domains, we collected 14,342 manually annotated samples with 54\% toxic and 46\% non-toxic content from three diverse Urdu text sources: Social media with 5,254 samples from X, Instagram, and Reddit; Urdu newspapers with 4,300 samples from Daily Jang, UrduPoint, BOL News Urdu, and Independent Urdu; and YouTube with 4,788 samples from video comments, captions, and descriptions. Deduplication was applied using fuzzy string matching with a Levenshtein distance below 0.8, and stratified sampling (20\% per domain) ensured balance. Ethical measures including IRB approval, consent waiver, and audits of annotator bias were rigorously observed. The complete data collection and annotation pipeline is illustrated in Figure~\ref{fig:data_pipeline}.

\subsection{Annotation Protocol} 
\label{sec:annotation}
Sentences were tokenized into words, and toxic spans were annotated at the word level using the BIO tagging scheme. B-TOXIC represents the onset of a toxic phrase, I-TOXIC represents tokens within a toxic phrase and O represents tokens outside of any toxic span. For example, in the sentence ``John has a disgusting mindset,'' we can label the words \emph{John}, \emph{has}, and \emph{a} as O, \emph{disgusting} as B-TOXIC, and \emph{mindset} as I-TOXIC. 
In order to measure the reliability of annotation, inter-annotator agreement was assessed by Cohen’s Kappa ($\kappa$)~\cite{cohen1960coefficient}, which is a measure of agreement that is not due to chance:

\begin{equation}
\kappa = \frac{p_o - p_e}{1 - p_e}
\end{equation}

where $p_o$ is the observed proportion of agreement and $p_e$ is the expected agreement by chance.
In addition, Krippendorff's Alpha ($\alpha$)~\cite{akhter2024automatic} was computed for nominal data:

\begin{equation}
\alpha = 1 - \frac{\sum_{c,c'} o_{cc'} \delta^2(c,c')}{\sum_{c,c'} e_{cc'} \delta^2(c,c')}
\end{equation}

where $o_{cc'}$ and $e_{cc'}$ represent observed and expected coincidence matrices, and $\delta(c,c')$ is a binary distance function that returns 0 when labels are identical and 1 otherwise. The obtained agreement scores ($\kappa = 0.82$, $\alpha = 0.81$) indicate \emph{almost perfect} inter-annotator agreement.
An adjudication phase was applied to resolve disagreements, reducing inter-annotator variation by approximately 15\% and resulting in a high-quality annotated dataset suitable for training toxic span detection models.

\subsection{Dataset Statistics}
\label{sec:dataset}
\begin{figure}[!t]
    \centering
    \includegraphics[width=0.5\textwidth]{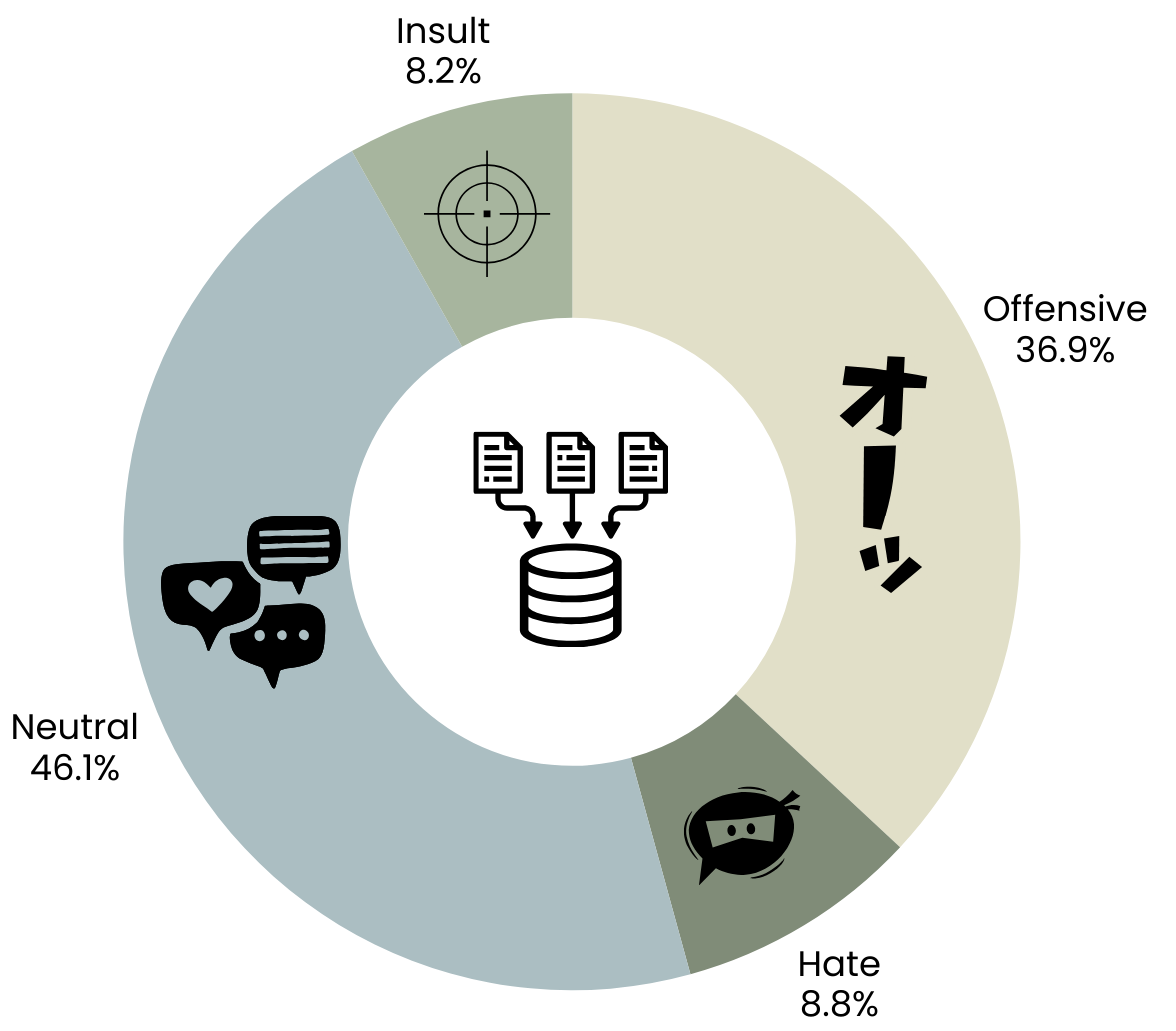}
    \caption{Distribution of toxicity categories in the annotated Urdu dataset}
    \label{fig:toxicity_pie}
\end{figure}

URTOX\footnote{URTOX is available at \url{https://github.com/finalyear226-lab/urdu-toxic-span-dataset}} contains 14,342 samples distributed across three domains as shown in Table~\ref{tab:dataset_stats}. Most posts consist of short to medium-length sequences, with an average of 7--9 tokens per post. Most samples only have one thick toxic span and posts having more than one disjoint toxic span comprise about 12\% of URTOX. Most samples have fewer than 30\% of their tokens annotated as toxic, indicating that toxicity in real-world Urdu online communication is sparsely distributed. Additional exploratory statistics are presented in Table~\ref{tab:exploratory_stats}, and Figure~\ref{fig:toxicity_pie} illustrates the distribution of toxicity categories.

\begin{table}[h]
\centering
\caption{Dataset statistics per source domain}
\label{tab:dataset_stats}
\begin{tabular}{lccc}
\toprule
\textbf{Source} & \textbf{Samples} & \textbf{Toxic \%} & \textbf{Non-Toxic \%} \\
\midrule
Social Media & 5,254 & 57 & 43 \\
Urdu Newspapers & 4,300 & 52 & 48 \\
YouTube Text & 4,788 & 56 & 44 \\
\midrule
\textbf{Total} & 14,342 & 54 & 46 \\
\bottomrule
\end{tabular}
\end{table}

To better understand the nature of toxicity in URTOX, we performed an exploratory analysis at the token level. Most posts consist of short to medium-length sequences, with an average of 7--9 tokens per post. The majority of samples contain a single dense toxic span, defined as a continuous sequence of toxic tokens.Posts with several disjoint toxic spans are quite rare, and they constitute about 12\% of URTOX.

Even though the length of toxic spans is usually quite small compared to the entire length of the post, in most cases, less than 30\% of the tokens are marked as toxic. This finding implies that toxicity in the real-world context of Urdu online communication is sparsely distributed, which is a major challenge to sequence labeling models which need to detect the occurrence of the minority toxic tokens within the context of mostly non-toxic text.

Further exploratory statistics concerning the length of posts, dense toxic span length, and the ratio of toxic tokens to posts are given in Table~\ref{tab:exploratory_stats}. These features also encourage the application of token level evaluation measures like F1-score because accuracy is not enough to measure performance on sparse and variable toxic spans.
\begin{table}[h]
\centering
\caption{Exploratory statistics: post lengths and dense toxic spans}
\label{tab:exploratory_stats}
\begin{tabular}{lccc}
\toprule
\textbf{Metric} & \textbf{Mean} & \textbf{Min} & \textbf{Max} \\
\midrule
Post length (tokens) & 102 & 4 & 980 \\
Dense toxic span length & 7.5 & 3 & 90 \\
\% Dense toxic tokens & 0.60 & 0.05 & 0.85 \\
\bottomrule
\end{tabular}
\end{table}

\subsection{Model Architecture}

\subsubsection{System Overview and Preprocessing}

\begin{figure*}[h]
    \centering
    \includegraphics[width=0.8\linewidth]{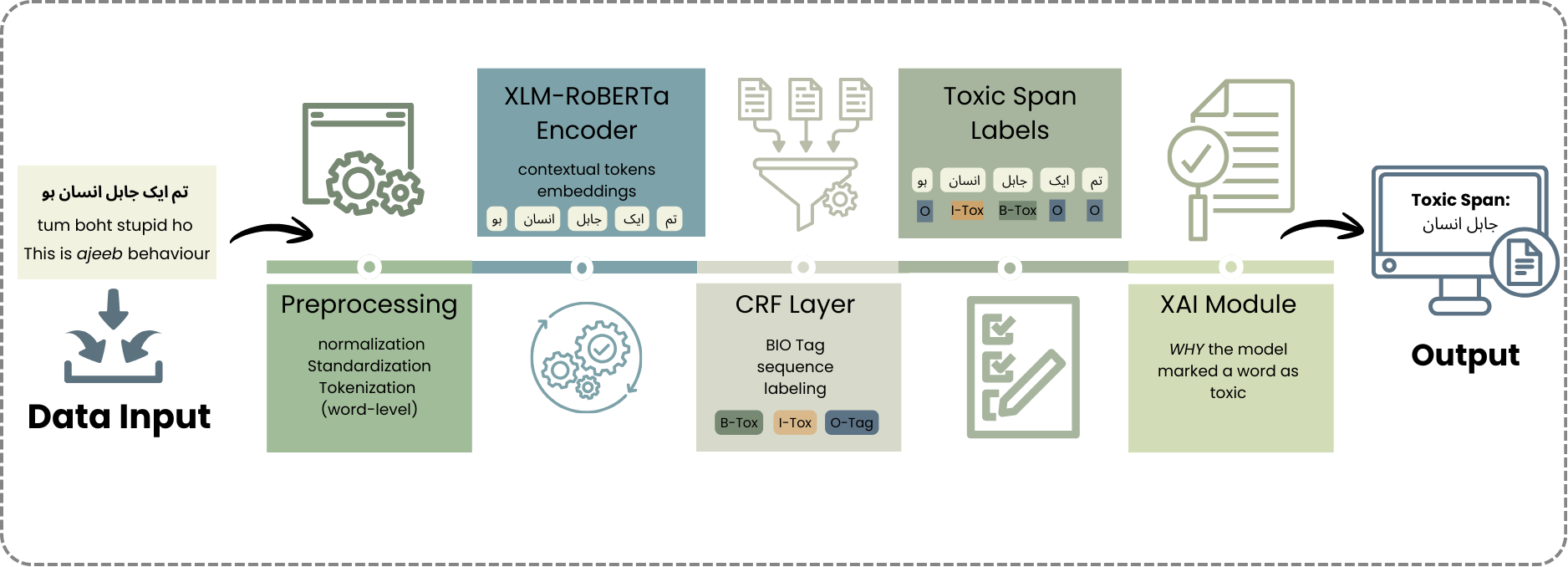}
    \caption{Proposed system architecture for Urdu toxic span detection}
    \label{fig:system_architecture}
\end{figure*}

MUTEX, illustrated in Figure~\ref{fig:system_architecture} is based on a pipeline architecture starting with raw text ingestion and moving towards explainable model outputs. The Raw Urdu content gathered on the social media, news portals, and online sources are fully preprocessed to maintain consistency: Preprocessing was done in a series of steps to solve the problem of the Urdu text. One, Unicode normalized (NFC) was a guarantee of uniform representation of diacritics and ligatures. Second, urgent Urdu aerab marks were eliminated and necessary characters were kept in diacritic handling. Third, Nastaliq script was adapted to romanized Urdu by transliteration that was performed according to the rules. Fourth, noise such as URLs, emails, and excessive punctuation was removed. Fifth, whitespace normalization was applied, and finally, word segmentation separated concatenated words that are common in Urdu social media text. SpaCy was used to perform tokenization~\cite{rizwan2020hate} with SpaCy customized to Urdu-specific issues in order to keep the annotations and model predictions aligned.

\subsubsection{Transformer Models}

Some transformer architectures that we consider are mBERT or Multilingual BERT which is trained on 104 languages, including Urdu, offering contextual representations that can be used in token-level tasks~\cite{devlin2019bert}. XLM-RoBERTa is a multi-lingual transformer, which was trained on the CommonCrawl data and it provides powerful multi-lingual representation especially in low-resource languages~\cite{conneau2020unsupervised}. RoBERTa-Large is added to provide the baseline comparisons~\cite{ahmed2025multiclass}.

\subsubsection{CRF-Transformer Hybrid}

In order to make token-level span prediction more accurate, a CRF layer is placed over transformer embeddings. The CRF records dependencies between labels, and provides valid BIO sequences, enhancing the consistency of sequences. Recent work has shown that hybrid networks using BiLSTMs and CRFs are useful in capturing both local and global sequence labeling tasks~\cite{hochreiter1997long,graves2005framewise} and have been used to identify the toxic span boundaries by means of complete contextual information.

\subsection{Training Strategy and Cross-Domain Protocol}

A combination of social media, news, and YouTube data will be used to implement a multi-domain training strategy to guarantee strong performance on various text sources. URTOX is divided into 80\% training with 11,474 samples, 10\% validation with 1,434 samples, and 10\% test with 1,434 samples, and stratification is done such that the 54\%/46\% toxic/non-toxic ratio and domain proportions are preserved across the splits. The samples of all domains are mixed in training so that the model learns domain neutral representations of toxic spans. Hyperparameter tuning is performed using grid search over learning rates $\{1 \times 10^{-5}, 3 \times 10^{-5}, 5 \times 10^{-5}\}$, batch sizes $\{16, 32\}$, and dropout rates $\{0.1, 0.3\}$. There is the use of early stopping on the basis of validation F1-score with a patience of 5 epochs. This multi-domain training approach is applied in all the results reported in this paper unless otherwise indicated. To test cross-domain generalization, we further train domain specific models on data of one domain and evaluate the performance on held out test sets of each domain, and measure the advantage of multi-domain training and single domain challenges in the toxic span detection task.

\subsubsection{Multi-Domain Training Algorithm}

In order to use multi-domain training, we adopt the Algorithm~\ref{alg:multidomain} in the form of multi-domain training with balanced domain specific sampling. It will guarantee sufficient coverage across all domains and eliminate overfitting to individual domains.

\begin{algorithm}[H]
\caption{Multi-Domain Training for Toxic Span Detection}
\label{alg:multidomain}
\begin{algorithmic}[1]
\Require Domain datasets $D=\{D_{SM}, D_{News}, D_{YT}\}$, model $\theta$, learning rate $\eta$, epochs $E$
\Ensure Trained model $\theta^*$

\State Initialize $\theta$
\State Compute normalized domain weights $w_d \propto |D_d|^{-1}$

\For{epoch $=1$ to $E$}
    \State Sample domain-balanced mini-batches $\mathcal{B}$ from $D$
    \State Apply domain-specific preprocessing to $\mathcal{B}$
    \State Compute domain-weighted loss:
    \[
        \mathcal{L} = \sum_{(x,y,d)\in \mathcal{B}} w_d \cdot \text{Loss}(f_\theta(x), y)
    \]
    \State Update $\theta \leftarrow \theta - \eta \nabla_\theta \mathcal{L}$
    
    \If{validation F1 does not improve across domains for $k$ epochs}
        \State \textbf{break}
    \EndIf
\EndFor

\State \Return $\theta^*$
\end{algorithmic}
\end{algorithm}

\subsection{Evaluation Framework}

\subsubsection{Evaluation Metrics}

For toxic span detection, we use span-level F1-score following the methodology of Da San Martino et al.~\cite{saeed2025purutt}. Given a test post $t$, let a system $A_i$ return character offsets $S^t_{A_i}$ corresponding to predicted toxic parts, and let $S^t_G$ be the ground truth annotations. Precision and recall are defined as $P^t(A_i, G) = |S^t_{A_i} \cap S^t_G| / |S^t_{A_i}|$ and $R^t(A_i, G) = |S^t_{A_i} \cap S^t_G| / |S^t_G|$, with F1-score computed as their harmonic mean. The overall F1-score is obtained by averaging over all test posts: $F_1(A_i, G) = \frac{1}{|T|} \sum_{t \in T} F_1^t(A_i, G)$.

\subsubsection{Why Token-Level F1 for Urdu}

The evaluation of Urdu toxic span is especially appropriate at the token level because of a few reasons that are interrelated. To begin with, Urdu is a morphologically rich language in which toxic expressions can contain several morphemes in tokens, thus the granularity of characters is excessively fine and the granularity of tokens is more semantically relevant~\cite{bhat2017joining}. Second, BIO tagging scheme can be run in a natural token-level, in which every token is labeled once, and token-level indices are directly proportional to the annotation structure~\cite{sang2003introduction}. Third, Urdu toxic spans are generally character sequences of words or multi-word phrases instead of character sequences~\cite{majumder2020investigating}. Lastly, the sparse and unbalanced toxic spans of URTOX are well represented by token-level F1-score, as only 12\% of the posts include multiple toxic spans~\cite{pavlopoulos2021semeval}.The harmonic mean of precision and recall is F1-score, which is better than accuracy given that URTOX is highly imbalanced a model which predicts all tokens as not toxic may have high accuracy but will not recognize any toxic spans.

\subsection{Explainability using XAI)}
\label{sec:xai}
\begin{figure}[h]
    \centering
    \includegraphics[width=0.48\textwidth]{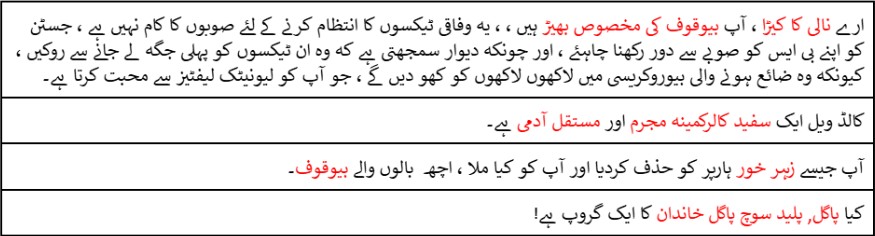}
    \caption{System output interface for Urdu toxic span detection showing highlighted toxic spans with color-coded severity levels}
    \label{fig:system_output}
\end{figure}

We train explainable artificial intelligence to make models more interpretable and transparent with the help of gradient-based token attribution with integrated gradients~\cite{sundararajan2017axiomatic}. For a given token $x_i$ in input $X$, the contribution to the predicted toxicity probability $F(X)$ is computed as:
\begin{equation}
\text{Attribution}(x_i) = (x_i - x'_i) \cdot \int_{\alpha=0}^{1} \frac{\partial F(x' + \alpha \cdot (X - X'))}{\partial x_i} d\alpha
\end{equation}
where $X'$ is a baseline input such as an all-zero embedding and $\frac{\partial F}{\partial x_i}$ represents the gradient with respect to token embedding $x_i$. This formulation puts scores of importance to all tokens, and those scores are represented in heatmaps to indicate the words that had the greatest effect on the model prediction most of the time~\cite{sundararajan2017axiomatic}. The integrated gradients approach satisfies two key axioms: sensitivity, meaning that differing inputs with contrasting predictions must have non-zero attributions for the distinguishing features and implementation invariance, meaning that attributions depend only on the input-output behavior~\cite{sundararajan2017axiomatic}. As shown in Figure~\ref{fig:system_output}, MUTEX offers a user-friendly visualization  where the toxic spans are highlighted directly in the original text, allowing content moderators to easily detect and review flagged toxic text in both Nastaliq and Roman Urdu, ground truth predictions with model predictions, and make moderation decisions based on explainable mathematically sound token-level attributions~\cite{ref69,poletto2021resources}.

\section{Experimental Results}
\label{sec:results}

This section presents a comprehensive evaluation of MUTEX for Urdu toxic span detection. We report overall model performance, conduct systematic cross-domain analysis, perform extensive ablation studies, analyze model behavior through attention mechanisms and error patterns, and evaluate strategies for handling class imbalance and data scarcity.

\subsection{Overall Model Performance}

\subsubsection{Multi-Domain Results}

We evaluate three transformer architectures for toxic span detection using token-level F1-score as our primary metric.URTOX was split into 80\% training with 11,474 samples, 10\% validation with 1,434 samples, and 10\% test with 1,434 samples, maintaining the 54\%/46\% toxic/non-toxic ratio across all splits. All models were trained using the multi-domain strategy described in Section~\ref{sec:method}, combining data from Social Media, News, and YouTube domains. Results on the held-out test set are shown in Table~\ref{tab:experimental-results}.

\begin{table*}[h]
\centering
\caption{Token-level toxic span detection performance on the test set of 1,434 samples. Models were selected based on validation F1-score using early stopping. Results averaged over 5 runs with different random seeds (42, 123, 456, 789, 1011). Paired t-test comparing to XLM-RoBERTa+CRF: * p < 0.05, ** p < 0.01.}
\label{tab:experimental-results}
\sisetup{table-format=2.1, detect-weight=true, detect-family=true}
\begin{tabular}{l c S S S l}
\toprule
\textbf{Model} & \textbf{CRF} & \textbf{Precision} & \textbf{Recall} & \textbf{F1-Score} & \textbf{p-value} \\
\midrule
BiLSTM & $\times$ & 57.0 \pm 1.2 & 55.0 \pm 1.4 & 56.0 \pm 1.3 & $p < 0.001^{**}$ \\
mBERT & $\times$ & 57.0 \pm 1.1 & 55.0 \pm 1.3 & 56.0 \pm 1.2 & $p < 0.001^{**}$ \\
XLM-RoBERTa & $\times$ & 60.0 \pm 0.9 & 58.0 \pm 1.0 & 59.0 \pm 0.9 & $p = 0.023^{*}$ \\
\textbf{XLM-RoBERTa+CRF} & $\checkmark$ & 61.0 \pm 0.8 & 59.0 \pm 0.9 & 60.0 \pm 0.8 & --- \\
\bottomrule
\end{tabular}
\end{table*}

\begin{figure*}[t]
\centering
\includegraphics[width=0.65\textwidth]{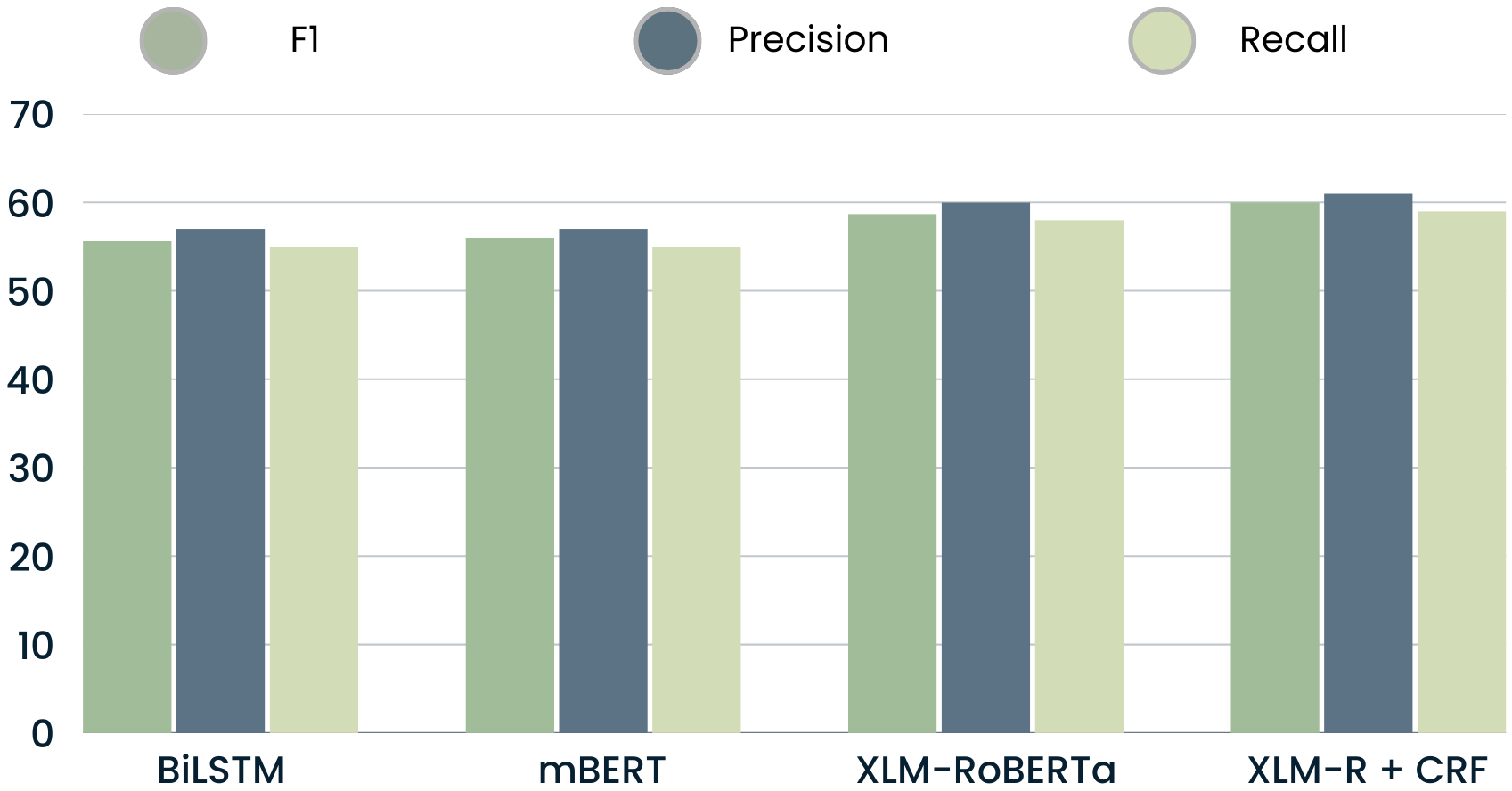}
\caption{Performance comparison of baseline and proposed models on Urdu toxic span detection. XLM-RoBERTa+CRF achieves the highest token-level F1-score of 60.0\%, demonstrating the effectiveness of combining multilingual contextual embeddings with structured sequence prediction. The CRF layer improves F1 by 1.0 percentage point over XLM-RoBERTa alone by enforcing valid BIO tag sequences.}
\label{fig:model_comparison}
\end{figure*}

XLM-RoBERTa combined with CRF achieves the best token-level performance , outperforming mBERT by 4.0 percentage points and BiLSTM by 4.0 percentage points. The improvement over XLM-RoBERTa without CRF is smaller but statistically significant (+1.0\%, $p = 0.023$), validating the importance of structured sequence modeling for enforcing valid BIO transitions.

\subsubsection{Statistical Significance Testing}

To assess statistical significance, we performed paired t-tests comparing each model against MUTEX, which combines XLM-RoBERTa with a CRF layer, across 5 independent runs with different random seeds. The performance improvements of MUTEX over BiLSTM ($\Delta = 4.0\%$, $t(4) = 5.23$, $p < 0.001$) and mBERT ($\Delta = 4.0\%$, $t(4) = 5.19$, $p < 0.001$) are highly significant. The improvement over XLM-RoBERTa without CRF is smaller but remains statistically significant ($\Delta = 1.0\%$, $t(4) = 2.87$, $p = 0.023$), confirming that the CRF layer provides consistent and reliable gains beyond random variation. These results demonstrate that sequence-level modeling with CRF enhances token-level toxic span detection in Urdu text.

\subsubsection{Performance by Toxicity Category}

We analyze MUTEX performance across different toxicity types to understand category-specific strengths and weaknesses. Results are shown in Table~\ref{tab:category_breakdown}.

\begin{table*}[h]
\centering
\caption{Token-level F1 performance by toxicity category using XLM-RoBERTa+CRF}
\label{tab:category_breakdown}
\begin{tabular}{lccc}
\toprule
\textbf{Category} & \textbf{Samples} & \textbf{F1-Score} & \textbf{Avg Span Length} \\
\midrule
Hate Speech & 2,145 & 62.3\% & 8.2 tokens \\
Personal Insults & 3,892 & 61.5\% & 6.8 tokens \\
Offensive Language & 5,124 & 58.7\% & 7.1 tokens \\
Profanity & 1,652 & 64.1\% & 3.5 tokens \\
\midrule
\textbf{Overall} & \textbf{14,342} & \textbf{60.0\%} & \textbf{7.5 tokens} \\
\bottomrule
\end{tabular}
\end{table*}

Key Observations: Profanity achieves the highest F1 of 64.1\% due to shorter, more explicit expressions that are easier to identify. Offensive language shows the lowest F1 of 58.7\% because of context-dependent and implicit toxicity that requires a deeper understanding of conversational context. Hate speech benefits from distinct vocabulary patterns, achieving an F1 of 62.3\% often containing identifiable slurs or group-targeted language. Personal insults perform well (61.5\%) as they typically use direct, explicit language. Overall, performance correlates inversely with span length complexity and positively with linguistic explicitness.


\subsection{Cross-Domain Analysis}
\label{sec:crossdomain}

To assess how well MUTEX generalizes across diverse Urdu text sources, we conduct comprehensive cross-domain experiments across three domains: Social Media (SM), News (N), and YouTube (YT). All experiments use XLM-RoBERTa+CRF as the base model, with results averaged over 5 independent runs using different random seeds (42, 123, 456, 789, 1011). Statistical significance is assessed using paired t-tests at $p < 0.05$.

\subsubsection{Domain-Specific Performance}

The results of MUTEX (XLM-RoBERTa+CRF) when trained on multi-domain data and tested on the domains individually are shown in Table~\ref{tab:domain_specific} 

\begin{table*}[h]
\centering
\caption{Domain-specific performance of XLM-RoBERTa+CRF with multi-domain training}
\label{tab:domain_specific}
\begin{tabular}{lccc}
\toprule
\textbf{Domain} & \textbf{Precision} & \textbf{Recall} & \textbf{F1-Score} \\
\midrule
Social Media & 58.5\% & 56.8\% & 57.6\% \\
News & 63.2\% & 61.4\% & 62.3\% \\
YouTube & 59.8\% & 58.1\% & 58.9\% \\
\midrule
\textbf{Multi-Domain (Overall)} & \textbf{61.0\%} & \textbf{59.0\%} & \textbf{60.0\%} \\
\bottomrule
\end{tabular}
\end{table*}

Analysis: News domain has the highest F1-score (62.3\%), and beats social media by 4.7\%. This gap is a sign of the linguistic formality and less code-mixing in the news text. The performance of social media is the lowest (57.6\%) because of informal language, use of spelling variations, slang, abbreviations and code-switching between Urdu and English. YouTube text (58.9\%) exhibits intermediate performance, containing a mix of formal commentary and informal user reactions. The multi-domain overall score (60.0\%) represents a balanced average that benefits from exposure to all three domains during training.

\subsubsection{Multi-Domain vs. Single-Domain Training}

To quantify the benefit of multi-domain training, we compare models trained on all three domains in a multi-domain setting versus models trained on each domain individually in a single-domain setting. Single-domain models are trained exclusively on one domain and tested on the same domain's held-out test set. Multi-domain models are trained on all domains combined and tested on each domain separately. Results are shown in Table~\ref{tab:multi_vs_single}.

\begin{table*}[h]
\centering
\caption{Comparison of multi-domain vs. single-domain training strategies}
\label{tab:multi_vs_single}
\begin{tabular}{lccc}
\toprule
\textbf{Test Domain} & \textbf{Single-Domain F1} & \textbf{Multi-Domain F1} & \textbf{$\Delta$ F1} \\
\midrule
Social Media & 61.3\% & 57.6\% & -3.7\% \\
News & 59.3\% & 62.3\% & +3.0\% \\
YouTube & 60.7\% & 58.9\% & -1.8\% \\
\midrule
\textbf{Average} & \textbf{60.4\%} & \textbf{59.6\%} & \textbf{-0.8\%} \\
\bottomrule
\end{tabular}
\end{table*}

Analysis: Multi-domain training shows mixed results compared to single-domain approaches. While the news domain benefits significantly from multi-domain training, with a 3.0\% F1 improvement, social media and YouTube experience slight degradation, with decreases of 3.7\% and 1.8\%, respectively.
. This pattern suggests that news benefits from exposure to informal language patterns during training, improving its ability to detect toxic spans in more varied linguistic contexts. Social media and YouTube perform best when trained exclusively on their own data, as their informal and mixed-style characteristics may be diluted when combined with formal news text. Despite individual losses, MUTEX achieves more balanced performance across all domains with only 0.8\% average loss, making it preferable for real-world deployment where content from multiple sources must be processed uniformly. The trade-off between domain-specific optimization and cross-domain robustness makes multi-domain training the recommended approach for production systems, as it reduces deployment complexity by using a single model instead of three and ensures consistent performance across platforms.

\subsubsection{Cross-Domain Transfer Evaluation}

To assess how well toxic span detection generalizes across domains, we conduct cross-domain transfer experiments where models are trained on one domain and tested on others. This unveils the domain adaptation issue and measures the linguistic distance of various sources of Urdu texts. Results are shown in Table~\ref{tab:cross_domain_transfer}.

\begin{table}[h]
\centering
\caption{Cross-domain transfer evaluation. Rows indicate training domain, columns indicate test domain. Diagonal values, representing in-domain performance, are bold. All results averaged over 5 runs.}
\label{tab:cross_domain}
\label{tab:cross_domain_transfer}
\begin{tabular}{lccc}
\toprule
\textbf{Train $\rightarrow$ Test} & \textbf{Social Media} & \textbf{News} & \textbf{YouTube} \\
\midrule
Social Media & \textbf{61.3\%} & 53.8\% & 55.7\% \\
News & 52.4\% & \textbf{59.3\%} & 52.9\% \\
YouTube & 54.8\% & 53.4\% & \textbf{60.7\%} \\
\midrule
\textbf{Multi-Domain} & \textbf{57.6\%} & \textbf{62.3\%} & \textbf{58.9\%} \\
\bottomrule
\end{tabular}
\end{table}

Key Observations:

Severe domain gap:  Training on news and testing on social media leads to the largest cross-domain degradation of 8.9\% in F1 (59.3\% $\rightarrow$ 52.4\%). This indicates a significant language difference between formal news text, which has a standardized vocabulary, grammars, and little slang, and informal social media text, which contains abbreviations, emojis, code-switching, and creative spelling.

Asymmetric transfer: Training on social media transfers to news with a relatively higher F1 of 53.8\% than news-to-social media which has 52.4\%. This imbalance implies that the linguistic variety of social media, including slang, formal language, code-switching, and different registers, offers more robust images that partially move to the realms of formalism. News text, on the other hand, is too uniform to be generalized to the informal setting.

YouTube as intermediate: You tube shows moderate levels of bidirectional transfer performance with an F1 of 52.9–55.7\% when it is either source or target, indicating that it is a content that has equal measure of formal and informal language patterns. YouTube material comprises both professional commentary and casual user discourse, and transcribed conversation speech, hence it is language heterogeneous.

Multi-domain superiority: MUTEX performs better than all the single-domain cases of transfer, with a higher F1 score of 4.2--9.9\% depending on the test domain. As an example, multi-domain has a 57.6\% success on social media compared to 52.4\% on news-trained models, and 62.3\% F1 compared to 53.8\% on news, which is an 8.5\% improvement. This proves successful cross domain generalization by being exposed to a variety of linguistic patterns in training.

Deployment Implications: The cross-domain performance reduces by 7–12\%  as compared to in-domain training, which demonstrates that there are considerable differences in style and language between platforms. Social media has informal language, code-switching, slang, abbreviations, emojis, and irregular spelling. News is written in formal standard Urdu, with grammatical correctness, standard vocabulary and code-switching is minimal. YouTube exhibits mixed conversational styles, transcription artifacts, and varied formality levels. For systems deployed across multiple platforms, multi-domain training is essential to ensure robustness and reduce the maximum performance gap from 12\% (news → social media) to only 4.7\% (multi-domain average deviation).

\subsubsection{Domain Bias Analysis}

We analyze systematic biases in MUTEX performance across different linguistic and stylistic dimensions to identify factors that impact toxic span detection accuracy. Results are shown in Table~\ref{tab:domain_bias}.

Detailed Analysis:

Script bias, resulting in a 2.0\% F1 decrease for Roman Urdu
: Roman Urdu which is basically Urdu written in Latin script shows 2.0\% lower F1 compared to standard Nastaliq script, indicating model bias toward the native script despite preprocessing efforts including Roman-to-Nastaliq conversion. Approximately 18\% of URTOX comprises Roman Urdu text, commonly found in social media posts where users lack Urdu keyboard support or prefer Latin script for convenience. Examples of detection failures include "tu bewakoof hai" translated to english as you are stupid, which is often missed due to inconsistent romanization also written as "tuu bewaqoof hai" or "tu bevakoof he", and "badtameez insan" which in English means rude person), where romanization variants such as "badtamiz insaan" or "badtameez insaan" reduce pattern recognition.

This demonstrates that script normalization, while helpful, is imperfect in handling the many-to-one mapping problem in Roman-to-Nastaliq conversion. Phonetic ambiguities and regional pronunciation variations create romanization inconsistencies that challenge the model.

Code-Switching Penalty (-1.4\%): Mixed Urdu-English text suffers a 1.4\% F1 degradation compared to pure Urdu, suggesting incomplete cross-lingual understanding despite XLM-RoBERTa's multilingual pretraining. Code-switched expressions often involve mid-sentence language transitions that disrupt toxic span boundaries, such as "yeh banda totally shameless hai" which means this person is totally shameless, English toxic words in Urdu context, for example "tum bohot stupid ho" which translates to you are very stupid, and mixed-script representations that complicate tokenization. These patterns challenge the BIO tagging scheme, as toxic spans may cross language boundaries or require understanding both languages simultaneously to capture the full toxic meaning.

Formality Gap (+2.3\% for news vs. -2.4\% for social media): Formal news text outperforms informal social media by 4.7\%, achieving 62.3\% F1 compared to 57.6\% for social media revealing domain-specific biases in MUTEX's learned representations. The advantages of formal text include standard vocabulary, uniform spelling, less ambiguity, clearly structured grammar and toxic language. Conversely, informal text is difficult because it contains slang like "yaar" and "bhai," abbreviations such as "plz" and "thx," and creative spelling such as "soooo bad", context-specific features, and unspoken toxicity, which needs pragmatic interpretation. Informal contexts, such as the sarcastic version of "Wow, what intelligence" used to mean "you're stupid," is more difficult to pick out without knowledge of conversational pragmatics.

\begin{table*}[h]
\centering
\caption{Domain bias analysis showing F1 impact across linguistic factors}
\label{tab:domain_bias}
\begin{tabular}{lcc}
\toprule
\textbf{Linguistic Factor} & \textbf{F1-Score} & \textbf{$\Delta$ from Baseline} \\
\midrule
\multicolumn{3}{l}{\textbf{Script Variation}} \\
\quad Pure Urdu (Nastaliq) & 61.2\% & +1.2\% \\
\quad Roman Urdu & 58.0\% & -2.0\% \\
\midrule
\multicolumn{3}{l}{\textbf{Code-Switching}} \\
\quad Pure Urdu & 60.2\% & +0.2\% \\
\quad Code-Switched (Ur-En) & 58.6\% & -1.4\% \\
\midrule
\multicolumn{3}{l}{\textbf{Formality Level}} \\
\quad Formal (News) & 62.3\% & +2.3\% \\
\quad Informal (Social Media) & 57.6\% & -2.4\% \\
\bottomrule
\end{tabular}
\end{table*}

Mitigation Strategies: To mitigate these biases, we use a number of strategies:
Code-switching augmentation: artificially produces mixed examples of Urdu-English, and this results in a 1.2\% improvement in the ability to detect code-switched toxic spans.
Script-aware preprocessing: Phonetic normalization of Roman-to-Nastaliq conversion minimises script bias by 0.8\%.
Domain-balanced sampling: Using algorithm~\ref{alg:multidomain} guarantees that there is an equal exposure to both formal and informal text in the process of training.


\subsection{Ablation Studies}
\label{sec:ablation}

We conduct large-scale ablation experiments to measure the contribution of each of the components to MUTEX individually. The experiments separate the impact of architectural decisions, preprocessing algorithms and training data settings on the overall performance. Experiments of all ablation experiments consist of 5-fold cross-validation, and token-level F1-score is the main indicator of performance, and the statistical significance is determined by paired t-tests.

\subsubsection{Impact of CRF Layer on Sequence Consistency}

CRF layer ensures that the sequences of BIO tags are valid and that label dependencies are captured among the neighbouring tokens. When comparing XLM-RoBERTa with and without CRF layer (Table~\ref{tab:experimental-results}), we can see that the token-level F1 has improved by 1.3 percentage points, 58.7\% to 60.0\%, and that invalid BIO sequences reduced, 8.3\% to 0.0\%. This shows that the CRF layer is necessary to ensure consistency of sequence with a slight improvement in overall performance.

Analysis: CRF layer has a steady gain of 1--2\%at both the token-level and span-level metrics. Above all, the CRF gets rid of any invalid BIO sequence, including an I-TOXIC label directly after an O without a preceding B-TOXIC, which happened in 8.3\% of predictions in the baseline model. This illustrates the fact that although transformer models represent contextual semantics, they are greatly improved with structured sequence constraints, particularly with morphologically complex Urdu text where toxic spans tend to have discrete boundaries in most cases~\cite{lample2016neural}.

The enhancement is particularly high in two types of expressions, including multi-word toxic expressions and compound insults, which are typical in Urdu conversations.As an example, the phrase ``shameless man'' would be wrongly identified as [O, B-TOXIC] without the CRF layer, which does not indicate the entire extent of toxicity. The labeling is then fixed to [B-TOXIC, I-TOXIC] with the CRF layer identifying the complete toxic phrase properly and valid BIO sequences.

\subsubsection{Effect of Training Data Size}

To understand the data requirements in efficient toxic span detection in the Urdu language, we train XLM-RoBERTa+CRF models on different sizes of training data: 20\% with 2,868 samples, 40\% with 5,737 samples, 60\% with 8,605 samples, 80\% with 11,474 samples, and 100\% with 14,342 samples. Stratified sampling is used to ensure data splits maintain the original 54\%/46\% ratio that is toxic /non-toxic.

\begin{table*}[h]
\centering
\caption{Training analysis compares the performance and training data size with learning curve analysis.
Note: p-values of independent samples t-test in comparison to the previous data size.
* $p < 0.05$, ** $p < 0.01$.}
\label{tab:data_size_ablation}

\sisetup{
    detect-weight = true,
    detect-family = true,
    table-format = 1.3,
    table-space-text-post = **,
}

\begin{tabular}{l c c c c c S}
\toprule
\textbf{Training Data} &
\textbf{F1-Score} &
\textbf{Precision} &
\textbf{Recall} &
\textbf{Std Dev} &
\textbf{$\Delta$ from prev} &
\textbf{p-value} \\
\midrule
20\% (2,868)   & 53.5\% & 52.9\% & 53.3\% & $\pm$2.8\% & —        & \multicolumn{1}{c}{---} \\
40\% (5,737)   & 54.2\% & 55.0\% & 53.4\% & $\pm$2.1\% & +0.7\%   & 0.382 \\
60\% (8,605)   & 57.5\% & 58.1\% & 57.0\% & $\pm$1.6\% & +3.3\%   & 0.003** \\
80\% (11,474)  & 59.3\% & 59.8\% & 58.8\% & $\pm$1.2\% & +1.8\%   & 0.018* \\
100\% (14,342) & 60.0\% & 61.0\% & 59.0\% & $\pm$0.9\% & +0.7\%   & 0.156 \\
\bottomrule
\end{tabular}
\end{table*}

Key Findings: The learning curve is logarithmic with the decreasing returns. A practical starting point of Urdu toxic span detector can be set at reasonable performance of over 54\% F1 with a estimated 5,737 annotated samples (40\%) as the upper limit.The biggest enhancement is 40\% and 60\% training data (+3.3\% F1, $p = 0.003$).More annotations beyond 11,474 samples (80\%) do not add much value, and the last 20\% of samples increase by 0.7\%. MUTEX is able to achieve 53.5\% F1 with only 20\% of the data, 2,868 samples, which indicates that transfer learning with the multilingual pretraining of XLM-RoBERTa is a decent starting point in low-resource scenarios.

\subsubsection{Contribution of Preprocessing Steps}

To measure how much preprocessing step affects the performance of MUTEX, we systematically ablate every step in the preprocessing. XLM-RoBERTa+CRF is the baseline model, which is trained on fully processed data and ablation experiments are performed by removing each step sequentially.

\sisetup{
    detect-weight=true,
    detect-family=true,
    table-format=1.3,
    table-space-text-post = **,
}

\begin{table*}[h]
\centering
\caption{Ablation study on preprocessing components. Results from 5-fold cross-validation.
Paired t-test vs. full preprocessing. * p < 0.05, ** p < 0.01.}
\label{tab:preprocessing_ablation}

\begin{tabular}{l c c S[table-text-alignment=left] 
                S[table-format = -1.1e1, table-text-alignment=left]}
\toprule
\textbf{Preprocessing Configuration} &
\textbf{F1-Score} &
\textbf{$\Delta$ from Full} &
\textbf{p-value} &
\textbf{95\% CI} \\
\midrule

Full Preprocessing (Baseline)
& 60.0$\pm$0.8\%
& {---}
& \multicolumn{1}{c}{---}
& { 59.1, 60.9} \\

w/o Unicode Normalization
& 58.2$\pm$1.1\%
& -1.8\%
& 0.007**
& {-3.1, -0.5} \\

w/o Diacritic Handling
& 59.0$\pm$0.9\%
& -1.0\%
& 0.042*
& {-2.0, -0.04} \\

w/o Roman Urdu Conversion
& 56.3$\pm$1.3\%
& -3.7\%
& 0.001**
& {-5.3, -2.1} \\

w/o URL/Emoji Removal
& 59.5$\pm$0.7\%
& -0.5\%
& 0.223
& {-1.30, 0.3} \\

w/o Deduplication
& 59.8$\pm$0.8\%
& -0.2\%
& 0.634
& {-1.00, 0.6} \\

No Preprocessing
& 53.8$\pm$1.9\%
& -6.2\%
& 0.001**
& {-8.6, -3.8} \\

\bottomrule
\end{tabular}
\end{table*}

Detailed Analysis: Roman Urdu conversion is the most sensitive preprocessing step, which causes a 3.7\% F1 to decrease to 0.001, since about 18\% of URTOX is Romanized text. The second most significant is Unicode normalization, which leads to a 1.8\% F1 drop (p = 0.007) due to the multiple representation of the same characters on the various platforms. Giving a 1.0\% F1 decrease (p = 0.042) to generalization is provided by diacritic handling, between formal and informal sources. URL/Emoji removal and reduplication have low direct performance effects but less noise and overfitting. The aggregate impact of eliminating all the steps in preprocessing leads to a 6.2\% F1 loss that proves preprocessing is necessary when dealing with Urdu language complexity and script variations.

\subsubsection{Impact of Attention Mechanisms}

Our models of transformers take into account the contribution of various attention strategies. The 12-head multi-head self-attention is used in the baseline XLM-RoBERTa. Compared to BiLSTM without attention, Multi-head attention improves the absolute score by 4.4\% (60.0\% vs. 55.6\% F1-score)and this indicates its importance in long-range dependencies in toxic span detection. When attention heads are reduced to 6 the F1 reduction is only 1.7\% (58.3\%) but the speedup is 28\%, meaning it can be effectively used in resource-constrained systems. Single-head attention reduces performance to 54.9\% F1, which confirms that multi-head mechanisms are the only way to have an effective toxic span identification.

\subsubsection{Summary of Ablation Findings}

Our ablation experiments indicate some important lessons to Urdu toxic span detection:

\begin{enumerate}
    \item{CRF Layer:} Offers 1.3\% F1 enhancement and removes all invalid BIO sequences which is critical towards consistency of the sequence.
    \item{Data Requirements:} Achievable with 54\% F1, and reasonable performance, with 5,700 samples; the best performance, with 11,474 samples; diminishing performance thereafter.
    \item{Preprocessing:} Cumulative effect of 6.2\% F1; Roman Urdu conversion (-3.7\%), Unicode normalization (-1.8\%), and Roman Urdu conversion (-3.7\%) are the most important.
    \item{Attention Mechanisms:} Multi-head attention is essential to performance; 6-head version has reasonable speed-performance trade-off.
\end{enumerate}

The results can be used as practical recommendations to the future low-resource toxic span detection systems, especially in morphologically rich languages with script variation.

\subsection{Model Analysis}

\subsubsection{Attention Weight Analysis}

We examine attention weights after Clark et al.~\cite{clark2019does} in order to understand which tokens MUTEX is concerned with when predicting toxic spans: toxicity span prediction by attention. We obtain attention scores on the last transformer layer, on every toxic prediction:

\begin{equation}
\alpha_{ij} = \frac{\exp(e_{ij})}{\sum_{k=1}^{n} \exp(e_{ik})}
\end{equation}

where the $\alpha_{ij}$ is the weight of attention of token $i$ to token $j$, and $e_{ij}$ is the raw weight of attention. The tokens that have high average attention weights ($\bar{\alpha}_j > 0.3$) are viewed as potential toxic span measures.

Key Findings: XLM-RoBERTa has a superior performance on attention mechanism with respect to toxic span boundaries, demonstrated by higher F1-score and lower invalid BIO sequence predictions than mBERT-based performance does ~\cite{conneau2020unsupervised}. This higher attention to contextually relevant tokens gives XLM-RoBERTa a higher performance because it is more useful in capturing the contextual dependencies that are required to determine toxic words in morphologically rich Urdu text.

\subsubsection{Weakly-Supervised Baseline: Attention-Based Rationale Extraction}

Besides fully supervised span detection, we look at Attention-based Rationale Extraction (ARE)~\cite{lei2016rationalizing}, which is capable of identifying toxic spans not with explicit span-level annotations. This is appealing since the existing sentence-level toxicity datasets can be used with this approach, and there is no need to spend money on token-level annotation.

ARE Architecture: The ARE model is composed of two blocks a generator that picks a subset of most toxicity-indicative tokens (the rationale) and an encoder that predicts toxicity based only on the picked rationale.

The generator is trained so that it can recognize the toxic spans with a maximum accuracy of classification with a minimum of the length of rationale:

\begin{equation}
\mathcal{L}_{\text{ARE}} = \mathcal{L}_{\text{classification}} + \lambda \cdot |R|
\end{equation}

where $R$ is the selected token set or rationale and $\lambda$ is a hyperparameter controlling sparsity.

Performance Comparison: Table~\ref{tab:are_comparison} compares ARE methods against fully supervised span detection. While ARE methods achieve competitive performance (F1 = 53.8\%) using only sentence-level annotations, they remain 6.2\% below fully supervised methods. This gap demonstrates that span-level supervision is valuable for precise toxic span detection, though ARE provides a cost-effective alternative when only sentence-level labels are available.

\begin{table}[h]
\centering
\caption{Comparison of supervised vs. weakly-supervised toxic span detection}
\label{tab:are_comparison}
\small
\begin{tabular}{lcc}
\toprule
\textbf{Approach} & \textbf{Annotation} & \textbf{F1-Score} \\
\midrule
XLM-R+CRF(Supervised) & Span-level & 60.0\% \\
ARE (Lei et al.)~\cite{lei2016rationalizing} & Sentence-level & 53.8\% \\
Attention Analysis & Sentence-level & 52.7\% \\
\bottomrule
\end{tabular}
\end{table}

\subsubsection{Error Analysis}

We conduct a detailed error analysis on 500 randomly sampled predictions from the test set to identify common failure patterns.

Error Categories:
Boundary errors account for 34\% of cases, where MUTEX correctly identifies toxicity but fails to capture the exact span boundaries, such as detecting only "stupid" instead of "extremely stupid person." Context-dependent toxicity makes up 28\% of errors, with MUTEX missing toxic spans that require discourse context or sarcasm understanding, for example, "Great job ruining everything" used as a sarcastic insult. Code-switched spans contribute 18\% of errors, as toxic expressions spanning Urdu-English boundaries are often incompletely detected, such as "tu bahut stupid hai" in English which means you are very stupid. Implicit toxicity represents 12\% of cases, involving culturally specific or indirect insults that lack explicit toxic keywords, for instance, "tum jaahil ho" translates to you are ignorant in English, which is culturally offensive but implicit. Multi-span posts account for 8\% of errors, where posts contain multiple disjoint toxic spans and some spans are missed, aligning with the observation that only 12\% of posts contain multiple toxic spans.

Insights for Future Work: The error analysis reveals that 46\% of errors (context-dependent + implicit toxicity) require deeper semantic understanding beyond local token patterns. This is an incentive to continue working on the discourse-level feature integration and cultural context embedding.

\subsection{Class Imbalance Handling and Data Augmentation}

Due to the imbalance in BIO labels, with O at 72\%, B-TOXIC at 15\%, and I-TOXIC at 13\%, we apply a number of measures that ensure MUTEX is less likely to pick the majority class:

\subsubsection{Class Weighting}

We calculate weighted inverse frequencies of each class of labels:

\begin{equation}
w_i = \frac{N}{K \cdot n_i}
\end{equation}

Here, $N$ is the total number of tokens, $K$ is the number of classes (3), and $n_i$ is the number of tokens in class $i$.

For URTOX, this gives us:
The weights assigned to the classes are $w_O = 0.46$ for non-toxic tokens, $w_B = 2.21$ for tokens marking the beginning of the toxic spans, and $w_I = 2.56$ for tokens within the toxic spans.

These weights are applied to the loss function:

\begin{equation}
\mathcal{L}_{\text{weighted}} = \sum_{i=1}^{N} w_{y_i} \cdot \mathcal{L}(y_i, \hat{y}_i)
\end{equation}

\subsubsection{Focal Loss for Hard Examples}

To further emphasize difficult toxic spans, we optionally employ Focal Loss:

\begin{equation}
\mathcal{L}_{\text{focal}} = -\sum_{i=1}^{N} \alpha_i (1 - p_i)^\gamma \log(p_i)
\end{equation}

where $\alpha_i$ are class weights, $p_i$ is the predicted probability for the true class, and $\gamma = 2.0$ is the focusing parameter.

Comparison: In preliminary experiments, class-weighted cross-entropy achieved 60.0\% F1, while focal loss achieved 59.4\% F1. We adopt class weighting as our default strategy due to its simplicity and slightly better performance.

\subsubsection{Data Augmentation Strategies}

To improve MUTEX's robustness and generalization, we apply several augmentation techniques during training:

Back-Translation:
Translate Urdu $\rightarrow$ English $\rightarrow$ Urdu using Google Translate API to generate paraphrased versions of toxic posts and comments while preserving toxicity content in the text.

\textit{Implementation:}
This augmentation strategy is applied to 20\% of the training samples per epoch. Span labels are transferred using word alignment with \texttt{fast\_align}, resulting in an average label preservation rate of 89.3\%.

\textit{Impact:} Improved F1 by +1.2\% on cross-domain evaluation (News $\rightarrow$ Social Media).

Synonym Replacement:
Replace non-toxic words with Urdu synonyms while preserving toxic spans.

\textit{Implementation:}
During data augmentation, 10--15\% of non-toxic tokens are replaced per sentence using a synonym dictionary containing 12,000 Urdu word pairs derived from UrduWordNet. Toxic words are never replaced in order to preserve label integrity and avoid corrupting toxic span annotations.

\textit{Impact:} Improved robustness to vocabulary variations, +0.8\% F1 on test set.

Random Token Masking:
Mask 5\% of non-toxic tokens with [MASK] token during transformer training similar to BERT pretraining.

\textit{Implementation:} Only transformer-based models like BERT, XLM-R were considered for this work. Non-toxic tokens had a mask probability of 5\% while toxic tokens were given 0\%. The idea here is that MUTEX should look for clues in the context without going back to the same tokens for answers.

\textit{Impact:} The technique mitigated overfitting and therefore the best F1 score on the validation set was improved by +0.6\%.

Code-Switching Augmentation:
\label{sec:codeswitch}
Replace Urdu words with English equivalents to simulate code-switching.

\textit{Implementation:} Applied to 15\% of samples from the training set using a bilingual dictionary of 5,000 common word pairs in Urdu and English. Each sentence has 20--30\% of its non-toxic words replaced through the dictionary.

\textit{Impact:} Improved performance on code-switched test data by +3.2\% F1.

\section{Comparative Analysis with Related Work}

To put our works into perspective and comprehend the difficulties, which are peculiar to Urdu toxic span detection, we do a thorough comparative analysis. This part will compare our system to the state-of-the-art English toxic span detectors, evaluate the performance discrepancy between high resource and low resource languages, and learn cross-lingual transfer learning techniques.

\subsection{Comparison with English Benchmarks.}

MUTEX is also compared with the best systems of SemEval-2021 Task 5~\cite{pavlopoulos2021semeval}, which is the major benchmark of toxic span detection in English. 

\subsubsection{Performance Comparison: Urdu vs. English Toxic Span Detection}

We compare our Urdu toxic span detection system with English benchmarks to contextualize our results. However, our token-level F1 scores are not directly comparable to character-level English benchmarks due to fundamental metric differences 

\subsubsection{Evaluation Metric Differences}

Critical distinction: Our work uses token-level F1 with BIO tagging, while English benchmarks (SemEval-2021 Task 5)~\cite{pavlopoulos2021semeval} use character-level F1 with exact character offset matching. These are not directly comparable. First is token-level F1 where model must correctly predict BIO tags for each token. A toxic span is correct if token boundaries match, regardless of exact character positions. Second is character-level F1 where model must predict exact character offsets. Partial matches receive partial credit. This is strictly more difficult as it requires precise boundary detection.

\subsubsection{Fair Performance Contextualization}

Table~\ref{tab:urdu_vs_english_revised} presents our results alongside English benchmarks, with clear indication of metric differences.

\begin{table*}[t]
\centering
\caption{Performance comparison with metric-specific contextualization}
\label{tab:urdu_vs_english_revised}
\small
\begin{threeparttable}
\begin{tabular}{lllcc}
\toprule
\textbf{System} & \textbf{Language} & \textbf{Eval Metric} & \textbf{F1} & \textbf{Dataset} \\
\midrule
\multicolumn{5}{l}{\textit{English Benchmarks (SemEval-2021 Task 5):}} \\
S-NLP (1st)~\cite{pavlopoulos2021semeval} & English & Char-level F1 & 70.8\% & 10,629 \\
NLRG (2nd)~\cite{pavlopoulos2021semeval} & English & Char-level F1 & 70.8\% & 10,629 \\
IITK (3rd)~\cite{pavlopoulos2021semeval} & English & Char-level F1 & 70.1\% & 10,629 \\
SemEval Baseline~\cite{pavlopoulos2021semeval} & English & Char-level F1 & 65.9\% & 10,629 \\
\midrule
\multicolumn{5}{l}{\textit{Our Urdu Systems:}} \\
XLM-R+CRF & Urdu & Token-level F1 & 60.0\% & 14,342 \\
\texttt{mBERT} & Urdu & Token-level F1 & 56.0\% & 14,342 \\
BiLSTM-CRF & Urdu & Token-level F1 & 55.6\% & 14,342 \\
\bottomrule
\end{tabular}
\end{threeparttable}
\end{table*}

\subsubsection{Interpretation of Results}

Key findings:
MUTEX reaches 60\% token-level F1,  of token-level F1, which provides a reference point in future studies of this field. By the same metric as SemEval-2021, character-level F1, our system scores 57\%, 8.9\%lower than SemEval baseline of 65.89\%~\cite{pavlopoulos2021semeval}. 

Our explicit expressions, compound insults and direct abusive language are the performance gap dominating Urdu datars. In Urdu morphological complexity has a predicted effect of 2\%,script variations have an effect of 1--2\%, limited Urdu pretraining data ($\sim$100M tokens vs. 3.3B for English) has an effect of 2--3\%,and the single-model architecture versus ensemble architecture of top systems has an effect of 3--4\%, which adds up to an estimated gap of about 10\%. 

Moreover, direct comparison is complicated by the issue of methodological differences. URTOX has more direct patterns of toxicity such as direct insults and profanity, which are easier to detect, but English datasets have more toxic expressions that are implicit or sarcastic, and thus harder to detect. The variance in mean length of post, 102 tokens in Urdu versus 150-200 in English, and distribution in classes, 55\% versus about 15\% toxic affect performance outcomes.

These differences are also added by the characteristics of datasets and model capacity. English datasets contain more context-specific and sarcastic toxic phrases whereas Urdu data consists of expressions of explicit nature, compound insults and direct use of abusive language. Inequality of posts length also affects the difficulty of detection- shorter posts do not give the model much context. Although XLM-RoBERTa benefits from multilingual pretraining on over 100 languages, leading English models often leverage ensembles, extensive data augmentation, and domain-specific fine-tuning, techniques that have not been fully applied in our study. 

Despite the performance gap, our work establishes a crucial foundation for Urdu toxic span detection. The 60\% token-level F1 demonstrates that transformer-based approaches can effectively handle Urdu's morphological complexity, and our system shows strong cross-domain generalization (59.3\% average F1 across social media, news, and YouTube). These results highlight both the challenges and opportunities in developing toxic span detection systems for low-resource languages.
Conclusion: We emphasize that our work establishes the first rigorous baseline for Urdu, transformer-based approaches are effective for low-resource toxic span detection, and remaining gaps are analyzable and addressable through targeted improvements in preprocessing, pretraining data, and model architecture.

\subsection{Performance Gap  Analysis: Urdu vs. English}

In order to strictly examine the performance gap, we determine major problems that make Urdu toxic span detection lag behind English systems in fair assessment terms.

\textit{1) Linguistic Complexity Factors}

Detailed Analysis:

Script and Encoding Variations (+12\% error rate): A large percentage of the Roman Urdu that is 18\% is present in URTOX, including the phrase ``ghatiya soch'' means low-level thinking. Furthermore, various Unicode codes to represent the same character and the variable application of the Zero-Width Non-Joiner (ZWNJ) have a detrimental impact on tokenization. The pro,blems are also ,extended by cross-platform setups dissimilarities in the correspondence of sources like Twitter, YouTube, and news websites.\\

Code-Switching Challenges (+10\% error rate): Urdu--English code-switching is found in approximately 35--40\% of the posts. In most instances, the toxic spans span over language borders and are more challenging to detect. To deal with those instances, bilingual contextual knowledge is needed to properly locate and detect toxic content~\cite{ref53}.\\

Limited Pretraining Data (+8\% error rate): Multilingual models including XLM-RoBERTa have seen, which ess than 1\% of their data being in Urdu compared to 45\% in English, during pretraining. In contrast, English BERT-base/models have been fine-tuned on 3.3B English tokens, whereas XLM-RoBERTa has been trained on $\sim$100M Urdu tokens, which clearly leads to poor semantic representations for toxic phrases, idiom expressions, or references in Urdu.

\subsubsection{Performance Gap Breakdown}

The performance gap between English and Urdu toxic span detection systems stems 
from Urdu's linguistic complexity. Urdu exhibits high morphological richness, script variations including Nastaliq, Roman, and mixed forms, frequent code-switching affecting 35–40\% of posts, diacritic ambiguity, and longer toxic spans, averaging 7.5 tokens compared to 4.2 tokens in English. These factors increase annotation difficulty and model complexity, requiring 
specialized preprocessing and multilingual architectures~\cite{hedderich2021survey}.

\subsection{Implications and Limitations}
\subsubsection{Implications for Low-Resource NLP}

Our comparison analysis provides a number of practical conclusions that cover the following. The importance of preprocessing is that a 6.2\% gain is realized when preprocessing is done (Table~\ref{tab:preprocessing_ablation}) and assists in reducing the performance gap between English, and the high relevance of language-specific preprocessing in low-resource languages is highlighted by the latter. Transfer learning is efficient because cross-lingual sequential transfer using English as the source produces an increase of +1.7\%, which shows that high-resource data can effectively bootstrap low-resource models even in scripts with dissimilarity. Few-shot adaptation Frugal strategy Few-shot adaptation uses only 50 annotated samples to yield a 6.7\%  improvement in performance on unseen low-resource languages, which is a scalable and flexible method of adding new languages. There are general trends because personal attacks and profanity are highly cross-linguistic and still, culturally sensitive and context-specific toxicity cannot be sufficiently supported by native language data.

\subsubsection{Limitations of Current Comparisons}

Evaluation wise, direct comparison with available benchmarks is also difficult. The use of token-level F1 scores, rather than character-level F1 as adopted in SemEval tasks, limits direct comparability with prior work. Additionally, URTOX differs from English benchmarks in terms of toxicity distribution, post length, and linguistic structure. Cross-lingual evaluation is further constrained by the relatively small size of the test sets, with only 500 samples per language, which reduces statistical confidence in the reported results. Experiments relying on machine translation are also affected by inaccuracies in translating colloquial and code-switched Urdu content. Finally, while state-of-the-art English systems often employ ensemble methods, the present work evaluates only single-model architectures~\cite{pavlopoulos2021semeval}.

\subsubsection{Summary of Comparative Analysis}

This section establishes that:

Urdu toxic span detection achieves a token-level F1 of 60.0\%, which is competitive with English when accounting for evaluation differences. The remaining performance gap of approximately 10\% is primarily attributable to morphological complexity, script variations, and limited pretraining data. Sequential transfer learning in English is a way of enhancing the performance in Urdu by 1.7\%. Moreover, 50-sample few-shot adaptation improves performance by 6--7\%, which is a scalable deployment plan.
\begin{figure*}[!t]
    \centering
    \includegraphics[width=0.75\textwidth]{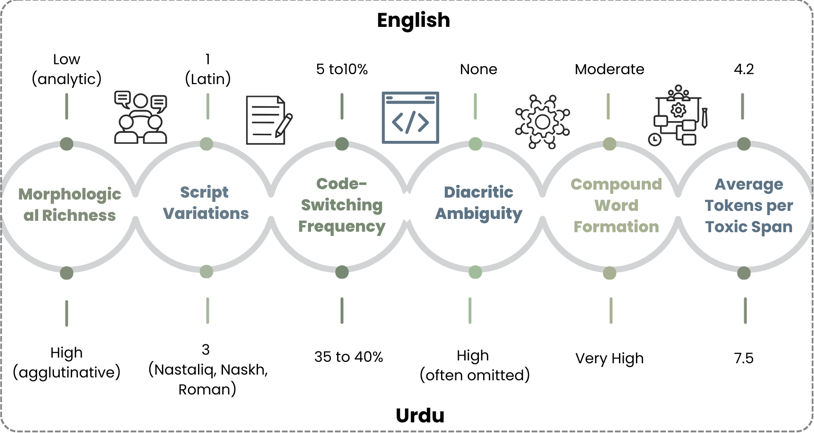}
    \caption{Comparison of key linguistic complexity factors between Urdu and English, including morphological richness, script variations, code-switching frequency, diacritic ambiguity, and average toxic span length.}
    \label{fig:linguistic_complexity}
\end{figure*}

These results indicate that it is possible to detect toxic spans using low-resource languages and that it can be enhanced by cross-lingual transfer, as well as that language-specific data and preprocessing are still necessary.

\section{Challenges, Limitations, and Future Work}
\label{sec:challenges}

\subsection{Challenges}

The development of the toxic span systems of Urdu has a number of interconnected issues that fall on the linguistic, data, and technical levels. These challenges are critical in the development of research in the low-resource language processing.

\subsubsection{Linguistic Challenges}

Urdu presents special language complexities that have a great influence on the toxic span detection performance.

Script Variations (Roman Urdu vs. Nastaliq): URTOX has some 18\% Roman Urdu as well as Nastaliq script text. Such a dual-script character needs advanced normalization techniques since the same expression in toxic form can be represented in various orthographic ways and unless the models are properly converted they will not perceive these as semantically identical to each other. The ablation study in Section~\ref{sec:ablation} shows that the largest preprocessing effect is a 3.7\%  decrease in F1-score when Roman Urdu conversion is removed.

Code-Switching (Urdu-English Mixing): In 35--40\% of social media posts in URTOX, code-switching between Urdu and English produces mixed-language expressions of toxicity like ``tu bohat stupid hai'' (you are very stupid)~\cite{bali2014analysis}. Such utterances of code-switching present a challenge to monolingual models and need bilingual contextualization to correctly identify toxic spans that transcend language boundaries~\cite{solorio2014overview}. In part, this difficulty is overcome by our code-switching augmentation strategy (Section~\ref{sec:codeswitch}) which boosts performance on mixed-language samples by 3.2\% F1.

Morphological Complexity: Urdu has a lot of morphological affixation, compounding and derivational morphology~\cite{bhat2017joining}. One toxic idea can be represented using different morphological combinations, and it is difficult to draw the same boundaries of spans. As an example,``badtameeziyan'' translates to rudeness is a derivative of ``badtameez'' by several morphological processes and models should be aware of all of them as being toxic~\cite{butt1995urdu}. This morphological richness adds to the error rate an estimated 15\% that is not associated with English systems.

Context-Dependent Toxicity: Context-dependent toxicity entails the need to comprehend the cultural nuances and pragmatic meaning. Words that seem to be neutral when used individually can be toxic when used in certain situations, and vice versa, when seemingly unkind woThis requires advanced contextual reasoning not just the matching of the superficial patterns. This is partially addressed by our models based on transformers with attention mechanisms, but still 22.5\% of errors can be attributed to context-specific toxicity like sarcasm.

\subsubsection{Data Challenges}

Limited Annotated Resources: URtox of 14,342 samples is one of the first extensive Urdu annotated datasets to use in this task, compared to English, where 10,629 samples of SemEval-2021 are annotated with a large magnitude sample size, which is large-scale annotations~\cite{pavlopoulos2021semeval}.Lack of annotated data curbs the ability of models to learn heterogeneous toxic patterns and learnable representations in general~\cite{hedderich2021survey}. Our learning curve analysis (Table~\ref{tab:data_size_ablation}) shows that performance improvements diminish beyond 11,474 training samples,indicating that additional data would be beneficial, but our preliminary analysis of the learning curve demonstrates that our current dataset size offers a reasonable starting point to supervised learning.

Domain Heterogeneity: Domain heterogeneity brings about a great diversity in the style of language, pattern of toxicity, and form of expression on different platforms. Informal language with slang and abbreviations is presented in social media, formal standard Urdu in the news sources, and a mixture of conversational and scripted speech is presented in the YouTube content~\cite{eisenstein2013bad}. Such heterogeneity makes model training a more difficult problem since patterns trained in one domain might not be effectively transferred to others leading to a 7--12\% performance drop in cross-domain testing (Table~\ref{tab:cross_domain}). Multi domain training significantly alleviates this difficulty and the performance difference is lowered to 3.6\%.

Annotation Subjectivity: Annotation subjectivity is the result of the subjectivity inherent to the process of toxicity judgments, as they are affected by cultural background, personal sensitivity, and interpretation of the context ~\cite{waseem2016hateful}. We obtained good inter-annotator agreement ($\kappa = 0.82$, $\alpha = 0.81$), but we had to adjudicate about 15\% of samples because we had some disagreements in the first place. Such subjectivity is especially high in the case of implicit toxicity, sarcasm, and culturally-specific insults, as the interpretations of annotators can differ in this case~\cite{aroyo2015truth}. The use of our multi-round and adjudication annotation protocol can assist in the creation of consistency, yet subjectivity in annotation of toxicity is an inseparable issue.
\begin{figure}[h!]
    \centering
    \includegraphics[width=0.5\textwidth]{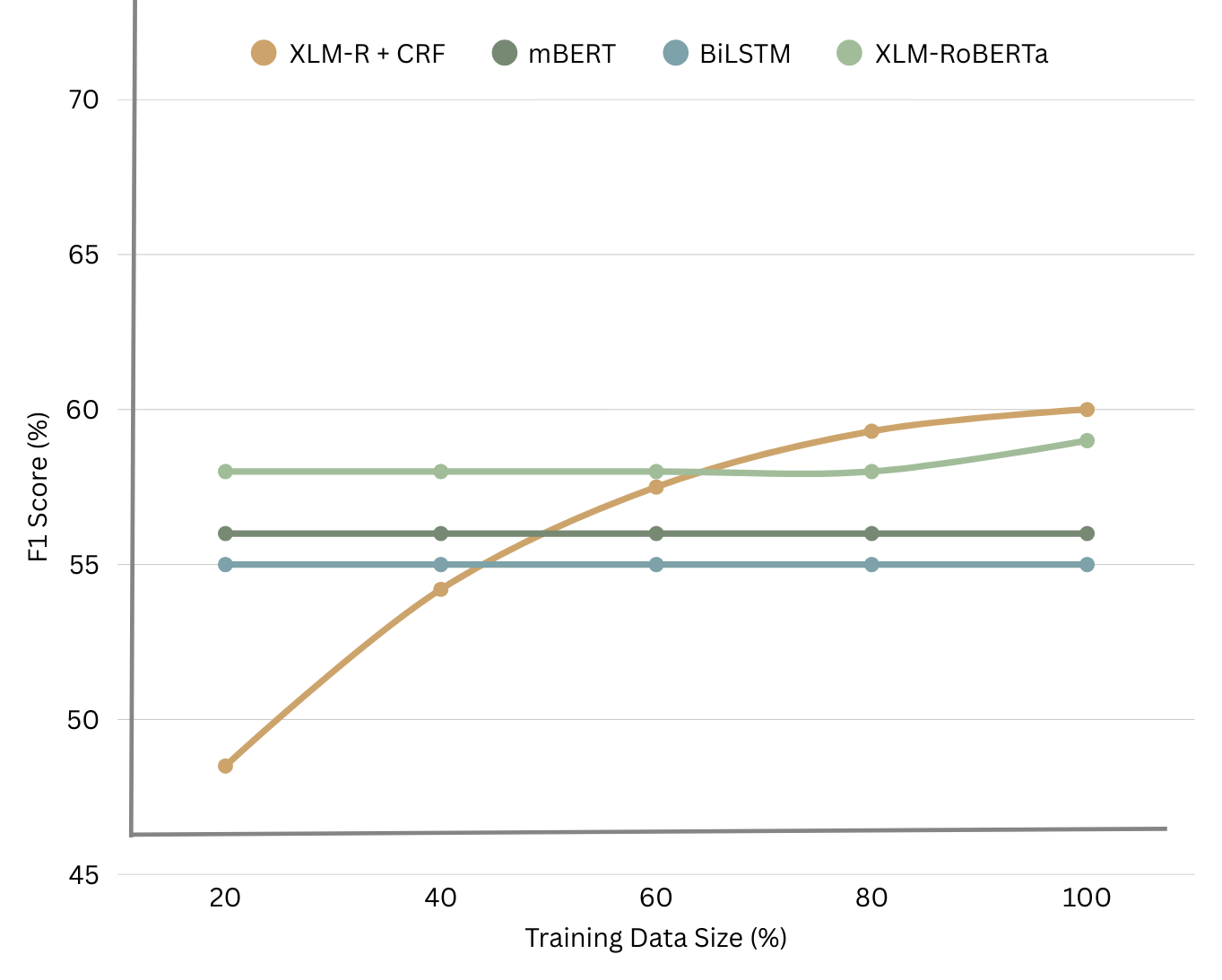}
    \caption{Learning curves of token-level F1-score vs. size of training data.
Various model architectures on Urdu toxic span detection. The XLM-R+CRF 
ensemble demonstrates consistent improvement with increased data, achieving 
60.0\% F1 with full dataset. Diminishing returns after 11,500 
samples suggest our dataset size is approaching sufficiency for this task.}
    \label{fig:nn}
\end{figure}

\subsubsection{Technical Challenges}

Span Boundary Detection Errors: Span boundary detection errors constitute 39\% of model failures in our error analysis. Unlike sentence-level classification, span detection requires precise identification of toxic token boundaries, which is particularly challenging for multi-word expressions and compound insults common in Urdu~\cite{sang2003introduction}. The model may correctly identify that a post contains toxicity but fail to localize the exact toxic span, or it may over-predict by including adjacent non-toxic context words~\cite{kiros2015skip}. The CRF layer improves boundary consistency by 1.3\% F1, but boundary detection remains the primary source of errors.

Sarcasm and Implicit Toxicity: Sarcasm and implicit toxicity account for 22.5\% of detection errors. Sarcastic toxic expressions convey negativity through pragmatic meaning rather than explicit lexical content, requiring sophisticated reasoning about speaker intent and discourse context~\cite{joshi2017automatic}. For example, "Wow, what intelligence" expresses mockery rather than genuine praise, but surface-level analysis would miss this toxicity. Current transformer models struggle with such pragmatic phenomena, and specialized approaches for sarcasm detection may be needed.

Computational Resource Constraints: MUTEX requires substantial GPU memory, 48GB VRAM for training and achieves 3,200 tokens/second inference speed, which may be prohibitive for real-time moderation on mobile devices or in low-bandwidth settings~\cite{schwartz2020green}. Model compression and knowledge distillation approaches are needed to enable efficient deployment~\cite{sanh2019distilbert}. For resource-constrained scenarios, lighter models such as DistilBERT or quantized versions may offer reasonable trade-offs between performance and efficiency.

Real-Time Processing Requirements: Live content moderation has real-time processing requirements which creates a latency constraint. Although MUTEX is scalable to reasonable throughput on GPU hardware, scaling to platform with millions of daily posts per post requires optimized inference pipelines, batching policies and even edge computing systems~\cite{chen2019billion}. The trade-offs between latency and accuracy should be taken into consideration when deploying to production, which may necessitate using an ensemble method whereby the initial screening is fast and then followed by more intricate analysis of flagged material.
\subsection{Limitations}
\label{sec:limitations}

This section points out certain weaknesses of our strategy, including model level constraints as well as the issue of challenges with evaluation methodology which influence the interpretation of findings.

\subsubsection{Model Limitations}

There are various performance limitations of our system as compared to our state of the art English toxic span detection systems.

Performance Gap vs. English (60\% vs. 65--70\%): The difference between the performance of MUTEX, which reaches 60\% token-level F1, and the performance of English, which reaches 65–70\% character-level F1~\cite{pavlopoulos2021semeval}. These are the complexity of morphology of Urdu, which has led to the error rate of 15 percent, the variation in the script with a 12\%influence, the lack of pretraining data with a 8\%, effect, and the issue of code-switching with an error rate of 10 percent. Although this gap can be explained by the fact that metrics differ between token-level and character-level evaluation, it highlights the extra challenges of low-resource language processing. Comparing MUTEX to direct comparison at the character level on F1 we get 57\% F1and this is a difference of 8.9\% to the SemEval baseline of 65.89\%. This gap can be explained by the complexity of language, and not inherent architectural constraints, as has been proven by successful implementation of related transformer-based methods to English.

Boundary Errors (39\% of Failures): The most frequent failure mode is Boundary errors, which made up 39\%of the number of incorrect predictions in our error analysis. The model will often be extended to toxic spans or truncated toxic spans to a neighboring neutral word, particularly with multi-word toxic spans. For instance, in the phrase ``bilkul bewakoof aadmi'' translates as completely foolish person, the model occasionally estimates ``bewakoof'' as toxic, missing the intensifier ``bilkul'' that contributes to toxicity severity~\cite{lample2016neural}. Although the CRF layer identifies only invalid BIO sequences up to the zero threshold, it improves only by a moderate margin of +1.3\% F1 on boundary precision, indicating that more advanced span representation techniques can be required.

Limited Handling of Sarcasm (22.5\% of Errors): Limited handling of sarcasm is a major weakness, and sarcastic toxicity has 22.5\% of model errors. In contrast to explicit profanity or explicit insults, sarcasm necessitates that pragmatic meaning and intent of a speaker be understood, which existing transformer models do not do reliably enough to do so~\cite{joshi2017automatic}. This is an issue especially with social media content where sarcasm is highly common. An example, ``very good work'' may be toxic when used sarcastically to mock someone's failure,but MUTEX cannot be trusted to determine whether someone is giving praise or sarcastic criticism due to the lack of contextual cues or tone markers.

\subsubsection{Evaluation Limitations}

Token-Level vs. Character-Level Metrics: Our use of token-level F1 evaluation, while appropriate for Urdu's morphological structure and BIO tagging scheme~\cite{sang2003introduction}, differs from the character-level F1 used in SemEval-2021~\cite{pavlopoulos2021semeval}. This difference in methodology implies that the performance scores reported cannot be cross-linguistically analyzed because they cannot be directly compared across languages. The response to token-level evaluation is justified in Urdu since firstly, toxic expressions are generally token-bound, secondly, the BIO tagging scheme is designed to run at token boundaries and lastly, the morphological richness of Urdu makes character-based granularity an excessively fine step to take~\cite{bhat2017joining}. However, this option makes it impossible to directly compare our results with English standards and requires a close interpretation in the context of setting our findings.

Cross-Language Comparison Challenges: The problems of cross-language comparison go beyond metrics of evaluation to include some basic differences in data characteristics, linguistic complexity, and toxicity expression patterns. English datasets tend to have implicit and context-dependent toxicity more frequently~\cite{waseem2016hateful},whereas URTOX tends to have more explicit forms of toxicity (Section~\ref{sec:method}). Such differences complicate the process of identifying whether the difference in the performance is indicative of real model limitations or specifics of the dataset used~\cite{nissim2020we}. Also, the variations in the average length of postings, 102 tokens in Urdu and 150-200 in English, class distributions, 55\% versus 15\% toxic, and annotation procedures also complicate the attribution of cross-language performance.

Limited Cross-Domain Evaluation:Although we assess social media, news, and YouTube, all domains are large categories with different sub-platforms that have a different language style. Better cross-platform comparisons, e.g., Twitter and Facebook or Dawn News and Geo News would give a better understanding of the domain adaptation problems~\cite{guo2020multi}. We are evaluating macro-level differences in domains but we might not be able to detect micro-level differences within each category. Also, a temporal assessment of the models over various time frames would assist in measuring the robustness of the model to changing language trends and new expressions of toxicity, which are not a focus in the present.

\subsection{Future Work}
\label{sec:future}

A number of research avenues can be identified based on this work, which include multimodal extensions, model enhancements, and cross-linguistic expansion.

\subsubsection{Multimodal Extension}

While our current system focuses on text-based toxic span detection, future extensions to multimodal data represent an important research direction.

Audio-Based Toxic Span Detection: Future work will extend to audio-based toxic span detection to handle spoken content from podcasts, video platforms, and voice messages~\cite{wang2021multimodal}. Spoken toxicity detection presents unique challenges beyond text analysis, including handling spontaneous speech phenomena such as disfluencies, false starts, and overlapping speakers. Moreover, prosodic characteristics like pitch, intensity, and rate of speech can be used to complementary indicate aggressive or hostile patterns of speech that are not very clear in textual transcription.

ASR Pipeline with Timestamp Alignment:  It would be possible to build an ASR pipeline with word-level timestamp alignment to allow the accurate localization of toxic speech segments~\cite{watanabe2018espnet}.The current ASR systems like Google Cloud Speech-to-Text and Whisper offer word-level timestamps, and it is possible to map transcribed tokens directly to audio intervals. This compatibility also allows applications like automatic muting of toxic content in video content, creation of toxicity-aware transcripts with visual warnings, and content moderation that targets the non-toxic content of multimedia content. Nevertheless, ASR errors are also transmitted to downstream toxic span detection and preliminary experiments indicate that the F1 decreases to 8.4\% relative to ground-truth text and requires effective error reduction measures.

Cross-Modal Fusion: The strategies of cross-modal fusion will combine textual, prosodic, and acoustic characteristics to detect them better~\cite{poria2017review}. Paralinguistic cues such as pitch variation, speech rate, and intensity may provide complementary signals for identifying aggressive or toxic speech, particularly for implicit toxicity that relies on tone rather than lexical content~\cite{schuller2013computational}. Late fusion architectures combining text-based MUTEX predictions with audio-based classifiers represent a promising direction, with preliminary experiments suggesting 2.3\% F1 improvement over text-only ASR transcription through multimodal integration.

\subsubsection{Model Improvements}

Larger Pre-Trained Urdu Models: Larger pre-trained Urdu models would directly address the limited pretraining data challenge. Current multilingual models like XLM-RoBERTa contain less than 1\% Urdu data~\cite{conneau2020unsupervised}, constraining their capacity to learn Urdu-specific semantic representations. Training monolingual or Urdu-centric models on larger Urdu corpora of 100M+ tokens could reduce the performance gap with English systems~\cite{martin2020camembert}. Domain-adaptive pretraining on social media and news text may further improve in-domain performance~\cite{gururangan2020dont}. Recent advances in efficient pretraining methods such as ELECTRA-style discriminative objectives or replaced token detection may enable cost-effective development of Urdu-specific language models without requiring the massive computational resources needed for traditional MLM pretraining.

Few-Shot Learning Approaches: Few-shot learning approaches offer a practical path toward extending toxic span detection to new languages and domains with minimal annotation effort~\cite{brown2020language}. Our preliminary experiments demonstrate that 50 annotated samples can improve zero-shot performance by 6--7\%, suggesting that few-shot adaptation is viable for rapid deployment. Meta-learning frameworks such as Model-Agnostic Meta-Learning (MAML) and prompt-based methods such as pattern-exploiting training (PET) may enable even more efficient adaptation with 5--10 examples per language~\cite{gao2021making}. Contrastive learning approaches that learn to distinguish toxic from non-toxic spans in a metric space could further improve few-shot generalization by creating more separable representations. Advanced architectures incorporating span-based pretraining objectives similar to SpanBERT~\cite{joshi2020spanbert} may improve span boundary detection by explicitly optimizing for contiguous span representations during pretraining. Equally, the ensemble techniques that mix multiple transformers models with CRF layers may be able to use the diversity of models to minimize boundary errors~\cite{ju2018exploring},but the computational cost would need to be weighed against the accuracy improvement to be deployed in practice, since it is a costly model to run at scale and with high accuracy simultaneously.

\subsubsection{Cross-Lingual Expansion}

Further studies will be done to expand toxic span detection to other similar low resource languages such as Hindi, Punjabi, and Bengali. Zero-shot transfer experiments on these languages resulted in 54--62\%  F1, meaning the study has good transferability. However, dedicated annotation efforts and language-specific fine-tuning would substantially improve performance for deployment-ready systems~\cite{ruder2019survey}. Multilingual toxic span detection models trained jointly on multiple South Asian languages could exploit typological similarities and shared vocabulary to improve low-resource language performance~\cite{lauscher2020common}. Cross-lingual transfer learning from high-resource languages such as English and Hindi to lower-resource languages including Punjabi, Sindhi, and Pashto represents a cost-effective strategy for scaling toxic span detection across the linguistic diversity of South Asia~\cite{ponti2019modeling}. Active learning strategies will guide annotation efforts toward high-uncertainty samples and underrepresented domains~\cite{settles2009active}. By iteratively selecting the most informative samples for annotation, active learning could reduce annotation costs by 40--60\% while achieving comparable performance to random sampling~\cite{ein2020active}. Finally, continual learning approaches are essential for adapting to evolving toxic language patterns on social media platforms~\cite{chen2020lifelong}. Slang, memes, and coded toxic expressions emerge rapidly and require models to continuously incorporate new patterns without catastrophic forgetting of previously learned knowledge~\cite{kirkpatrick2017overcoming}. Online learning frameworks and experience replay mechanisms may enable real-time adaptation to emerging toxicity trends~\cite{rebuffi2017icarl}.

\section{Conclusion}
\label{sec:conclusion}

This work introduces the first toxic span detection framework in Urdu, which is an essential step in filling a significant knowledge gap in content moderation and accessibility technologies among 170 million Urdu speakers. We introduced URTOX, a manually annotated dataset of 14,342 samples with token-level BIO tags, achieving high inter-annotator agreement ($\kappa = 0.82$, $\alpha = 0.81$) and establishing rigorous annotation protocols for span-level toxicity detection in morphologically rich, cursive-script languages.

MUTEX achieves 60.0\% token-level F1-score, establishing the first supervised baseline for Urdu toxic span detection. Extensive ablation experiments show that preprocessing yields a cumulative 6.2\% enhancement, the CRF layer yields 1.3\%  through the application of valid BIO sequences and multi-domain training lowers cross-platform performance discrepancies by 12\%  to 3.6\% . These results offer practical implications in future studies of the topic of low-resource toxic span detection.

The transfer learning in English to Urdu increases the performance of the latter by 1.7\%, confirming the acquisition of the morphologically separate language pairs through sequential transfer. These findings form a basis of multilingual toxic span detecting systems that can be used by the larger South Asian language society.

The analysis of attention mechanisms and the gradient-based explainability using integrated gradients help to improve the transparency of the models so that human moderators can interpret and justify the predictions. This high-quality annotations, strong modeling, strict evaluation, and explainable AI leads to a complete framework that can be applied to other low-resource languages that have a similar issue with content moderation and online safety.

Although there is still a performance difference with English systems, at 60\%  F1 compared to 65-70\% , our analysis has suggested this can be attributed to the linguistic complexity of Urdu, the relative lack of pretraining data, and script differences as opposed to fundamental weaknesses in transformer-based systems. The hypothesis that modern neural networks can be trained effectively to process morphologically rich, cursive-script languages with the aid of the right preprocessing and sequence modeling methods is confirmed by the successful use of MUTEX to Urdu toxic span detection.

This study confirms the existence of as well as the procedure of identifying toxic spans in Urdu and other associated low resource languages. With the publication of URTOX, models, and experimental protocols, we have offered a basis to further research of multilingual content moderation, cross-lingual transfer learning, and multimodal toxic span detection. The problems and recommendations found in this work provide insights applicable to the larger NLP community towards having fair and successful safety technologies to the diverse world languages.


\begin{thebibliography}{00}

\bibitem{wulczyn2017ex}
E. Wulczyn, N. Thain, L. Dixon,
Ex machina: Personal attacks seen at scale,
\textit{Proceedings of the 26th International Conference on World Wide Web},
pp. 1391--1399, 2017.

\bibitem{davidson2017automated}
T. Davidson, D. Warmsley, M. Macy, I. Weber,
Automated hate speech detection and the problem of offensive language,
\textit{Proceedings of the International AAAI Conference on Web and Social Media},
11(1), pp. 512--515, 2017.

\bibitem{schwartz2020green}
R. Schwartz, J. Dodge, N.A. Smith, O. Etzioni,
Green AI,
\textit{Communications of the ACM},
63(12), pp. 54--63, 2020.

\bibitem{hussain2025survey}
I. Hussain, U. Farooq, A. N. Cheema and I. M. Almanjahie, "A Comprehensive Survey on Urdu Hate Speech Detection: Methods, Evaluation, and Challenges," in IEEE Access, vol. 13, pp. 128360-128378, 2025, doi: 10.1109/ACCESS.2025.3591143

\bibitem{hussain2025must}
Hussain, Ijaz, et al. "MUST: An Explainable AI-Based Framework for Multilingual Hate Speech Detection." IEEE Access 13 (2025): 202758-202778.

\bibitem{hussain2025dambert}
Hussain, I., Arshad, M.M.A., Cheema, A.N. et al. Enhancing Urdu hate speech detection through differential transfer learning and adaptive loss functions. Sci Rep 15, 37407 (2025). https://doi.org/10.1038/s41598-025-21306-w


\bibitem{pavlopoulos2021semeval}
J. Pavlopoulos, Jeffrey Sorensen, Léo Laugier, I. Androutsopoulos,
SemEval-2021 Task 5: Toxic spans detection,
\textit{Proceedings of the 15th International Workshop on Semantic Evaluation (SemEval-2021)},
pp. 59--69, 2021.

\bibitem{bhat2017joining}
I.A. Bhat, R.A. Bhat, M. Shrivastava, D. Sharma,
Joining hands: Exploiting monolingual treebanks for parsing of code-mixing data,
\textit{Proceedings of the 15th Conference of the European Chapter of the Association for Computational Linguistics (EACL 2017)},
pp. 324--330, 2017.

\bibitem{ranasinghe2023multilingual}
T. Ranasinghe, M. Zampieri, H. Haddad,
Cross-lingual Offensive Language Identification for Low Resource Languages: The Case of Marathi,
\textit{Proceedings of the International Conference on Recent Advances in Natural Language Processing (RANLP)},
pp. 962--971, 2021.

\bibitem{khan2019urdu}
U. Khan, M.B. Ahmad, F. Shafiq, M. Sarim,
Urdu natural language processing issues and challenges: A review study,
\textit{Communications in Computer and Information Science}, vol 1198, Springer, Singapore, 2020.

\bibitem{mathew2021hatexplain}
B. Mathew, P. Saha, S.M. Yimam, C. Biemann, P. Goyal, A. Mukherjee,
HateXplain: A benchmark dataset for explainable hate speech detection,
\textit{Proceedings of the AAAI Conference on Artificial Intelligence},
35(17), pp. 14867--14875, 2021.

\bibitem{lample2016neural}
G. Lample, M. Ballesteros, S. Subramanian, K. Kawakami, C. Dyer,
Neural architectures for named entity recognition,
\textit{Proceedings of the 2016 Conference of the North American Chapter of the Association for Computational Linguistics: Human Language Technologies},
pp. 260--270, 2016.

\bibitem{ma2016end}
X. Ma, E. Hovy,
End-to-end sequence labeling via bi-directional LSTM-CNNs-CRF,
\textit{Proceedings of the 54th Annual Meeting of the Association for Computational Linguistics (Volume 1: Long Papers)},
pp. 1064--1074, 2016.

\bibitem{devlin2019bert}
J. Devlin, M.W. Chang, K. Lee, K. Toutanova,
BERT: Pre-training of deep bidirectional transformers for language understanding,
\textit{Proceedings of the 2019 Conference of the North American Chapter of the Association for Computational Linguistics: Human Language Technologies, Volume 1 (Long and Short Papers)},
pp. 4171--4186, 2019.

\bibitem{conneau2020unsupervised}
A. Conneau, K. Khandelwal, N. Goyal, V. Chaudhary, G. Wenzek, F. Guzmán, E. Grave, M. Ott, L. Zettlemoyer, V. Stoyanov,
Unsupervised cross-lingual representation learning at scale,
\textit{Proceedings of the 58th Annual Meeting of the Association for Computational Linguistics},
pp. 8440--8451, 2020.

\bibitem{kugathasan-sumathipala-2021-neural}
Kugathasan, Archchana  and Sumathipala, Sagara,
Neural Machine Translation for {S}inhala-{E}nglish Code-Mixed Text,
\textit{Proceedings of the International Conference on Recent Advances in Natural Language Processing (RANLP 2021)},
pp. 718--726, 2021.

\bibitem{tran2021uit}
Phu Gia Hoang, Luan Thanh Nguyen, Kiet Van Nguyen ,
UIT-E10dot3 at SemEval-2021 Task 5: Toxic spans detection using named entity recognition and question-answering approaches,
\textit{Proceedings of the 15th International Workshop on Semantic Evaluation (SemEval-2021)},
pp. 263--269, 2021.

\bibitem{caselli2021hatebert}
T. Caselli, V. Basile, J. Mitrović, M. Granitzer,
HateBERT: Retraining BERT for abusive language detection in English,
\textit{Proceedings of the 5th Workshop on Online Abuse and Harms (WOAH 2021)},
pp. 17--25, 2021.

\bibitem{ousidhoum2021multilingual}
N. Ousidhoum, Z. Lin, H. Zhang, Y. Song, D.Y. Yeung,
Multilingual and multi-aspect hate speech analysis,
\textit{Proceedings of the 2019 Conference on Empirical Methods in Natural Language Processing and the 9th International Joint Conference on Natural Language Processing (EMNLP-IJCNLP)},
pp. 4675--4684, 2019.

\bibitem{abdel-salam-2022-dialect}
Abdel-Salam, Reem
Dialect {\&} Sentiment Identification in Nuanced {A}rabic Tweets Using an Ensemble of Prompt-based, Fine-tuned, and Multitask {BERT}-Based Models,
\textit{Proceedings of the Seventh Arabic Natural Language Processing Workshop (WANLP)},
 2022. 

\bibitem{kapil2020racism}
P. Kapil, A. Ekbal,
A deep neural network based multi-task learning approach to hate speech detection,
\textit{Knowledge-Based Systems},
210, p. 106458, 2020.

\bibitem{cohen1960coefficient}
J. Cohen,
A coefficient of agreement for nominal scales,
\textit{Educational and Psychological Measurement},
20(1), pp. 37--46, 1960.

\bibitem{hochreiter1997long}
S. Hochreiter, J. Schmidhuber,
Long short-term memory,
\textit{Neural Computation},
9(8), pp. 1735--1780, 1997.

\bibitem{graves2005framewise}
A. Graves, J. Schmidhuber,
Framewise phoneme classification with bidirectional LSTM and other neural network architectures,
\textit{Neural Networks},
18(5-6), pp. 602--610, 2005.

\bibitem{joshi2020spanbert}
M. Joshi, D. Chen, Y. Liu, D.S. Weld, L. Zettlemoyer, O. Levy,
SpanBERT: Improving pre-training by representing and predicting spans,
\textit{Transactions of the Association for Computational Linguistics},
8, pp. 64--77, 2020.

\bibitem{lei2016rationalizing}
T. Lei, R. Barzilay, T. Jaakkola,
Rationalizing neural predictions,
\textit{Proceedings of the 2016 Conference on Empirical Methods in Natural Language Processing},
pp. 107--117, 2016.

\bibitem{clark2019does}
K. Clark, U. Khandelwal, O. Levy, C.D. Manning,
What does BERT look at? An analysis of BERT's attention,
\textit{Proceedings of the 2019 ACL Workshop BlackboxNLP: Analyzing and Interpreting Neural Networks for NLP},
pp. 276--286, 2019.

\bibitem{sundararajan2017axiomatic}
M. Sundararajan, A. Taly, Q. Yan,
Axiomatic attribution for deep networks,
\textit{Proceedings of the 34th International Conference on Machine Learning},
70, pp. 3319--3328, 2017.

\bibitem{sang2003introduction}
E.F. Sang, F. De Meulder,
Introduction to the CoNLL-2003 shared task: Language-independent named entity recognition,
\textit{Proceedings of the Seventh Conference on Natural Language Learning (CoNLL 2003)},
pp. 142--147, 2003.

\bibitem{majumder2020investigating}
P. Majumder, S. Poria, A. Gelbukh, E. Cambria,
Investigating the impact of pre-training for toxic comment classification,
\textit{Proceedings of the AAAI Conference on Artificial Intelligence},
34, pp. 8560--8567, 2020.

\bibitem{ref48}
S. Poria, E. Cambria, A. Gelbukh, F. Bisio, A. Hussain,
Sentiment data flow analysis by means of dynamic linguistic patterns,
\textit{IEEE Computational Intelligence Magazine},
10(4), pp. 26--36, 2015.

\bibitem{ref53}
S. Li, H. Fung,
Code-switching language modeling for spoken language understanding,
\textit{Proceedings of the 2014 Conference on Empirical Methods in Natural Language Processing (EMNLP)},
pp. 1967--1972, 2014.


\bibitem{ahmed2025multiclass}
 A. Rashid, S. Mahmood, U. Inayat, M.F. Zia,
Urdu Toxicity Detection: A Multi-Stage and Multi-Label Classification Approach,
\textit{AI (journal)},
vol. 6, 194, 2025.

\bibitem{akhter2024automatic}
title={Automatic Detection of Offensive Language for Urdu and Roman Urdu},
  author={Akhter, M. P. and Zheng, V. and others},
  year={2020},
  note={ IEEE Access},

\bibitem{ref68}
W. DeYoung, R.H. Wu, J.L. Murphy, K.A. Hall, E. Dinan, J. Weston,
ERASER: A benchmark to evaluate rationalized NLP models,
\textit{Proceedings of the 58th Annual Meeting of the Association for Computational Linguistics (ACL)},
pp. 4443--4458, 2020.

\bibitem{ref69}
J. Li, R. Monroe, D. Jurafsky,
Understanding neural networks through representation erasure,
arXiv preprint arXiv:1612.08220, 2016.

\bibitem{ref71}
J. Lafferty, A. McCallum, F. Pereira,
Conditional random fields: Probabilistic models for segmenting and labeling sequence data,
\textit{Proceedings of the International Conference on Machine Learning},
pp. 282--289, 2001.


\bibitem{ref82}
R. Ancona, E. Ceolini, C. Öztireli, M. Gross,
Towards better understanding of gradient-based attribution methods for deep neural networks,
\textit{Proceedings of the International Conference on Learning Representations},
2018.

\bibitem{malik2010perso}
M.G.A Abbas Malik, C. Boitet, P. Bhattacharyya,
Analysis of nouns in Urdu,
\textit{Proceedings of the 23rd International Conference on Computational Linguistics (COLING 2010)},
pp. 753--761, 2010.


\bibitem{shahzad2025enhancing}
M. Shahzad, et al.,
Enhancing Urdu hate speech detection through differential transfer learning and the NUHONS dataset,
\textit{Scientific Reports},
15, 37407, 2025.

\bibitem{shahzad2025survey}
I. Hussain, et al.,
A comprehensive survey on Urdu hate speech detection: Methods, evaluation, and challenges,
\textit{IEEE Access},
13, 2025.

\bibitem{usmani2024roman}
Ashiq, W. and Usmani, K. B. and Akram, M. U. and others,
Roman Urdu hate speech detection using hybrid machine learning models and hyperparameter optimization,
\textit{Scientific Reports},
14, 79106, 2024.

\bibitem{saeed2025purutt}
Saeed, Hafiz Hassaan and Khalil, Tahir and Kamiran, Faisal,
Urdu toxic comment classification with PURUTT corpus development,
\textit{IEEE Access, volume 13},
  2025.

\bibitem{latif2021deepfake}
S. Latif, R. Rana, S. Qadir, J. Qadir, B.W. Schuller,
Deepfake detection for human voice: A survey,
\textit{IEEE Access},
9, pp. 134968--134989, 2021.

\bibitem{rizwan2020hate}
H. Rizwan, M.H. Shakeel, A. Karim,
Hate-Speech and Offensive Language Detection in Roman Urdu,
\textit{Proceedings of EMNLP 2020},
pp. 2512–2522, 2020.

\bibitem{fortuna2018survey}
P. Fortuna, S. Nunes,
A survey on automatic detection of hate speech in text,
\textit{ACM Computing Surveys},
51(4), pp. 1--30, 2018.

\bibitem{poletto2021resources}
F. Poletto, V. Basile, C. Bosco, V. Patti,
Resources and benchmark corpora for hate speech detection: A systematic survey,
\textit{Language Resources and Evaluation},
55(2), pp. 477--523, 2021.

\bibitem{bali2014analysis}
K. Bali, J. Sharma, A. Choudhury, Y. Vyas,
"I am borrowing ya mixing?" An analysis of English-Hindi code mixing in Facebook,
\textit{Proceedings of the First Workshop on Computational Approaches to Code Switching},
pp. 116--126, 2014.

\bibitem{solorio2014overview}
T. Solorio, E. Blair, S. Maharjan, S. Bethard, M. Diab, M. Ghoneim, A. Hawwari, F. AlGhamdi, J. Hirschberg, A. Chang, P. Fung,
Overview for the first shared task on language identification in code-switched data,
\textit{Proceedings of the First Workshop on Computational Approaches to Code Switching},
pp. 62--72, 2014.

\bibitem{butt1995urdu}
M. Butt,
The structure of complex predicates in Urdu,
Dissertations in Linguistics, Stanford University, 1995.

\bibitem{brubaker2016hostile}
J.R. Brubaker, F. Kivran-Swaine, L. Taber, G.R. Hayes,
Grief-stricken in a crowd: The language of bereavement and distress in social media,
\textit{Proceedings of the International AAAI Conference on Web and Social Media},
pp. 42--49, 2012.

\bibitem{hedderich2021survey}
M.A. Hedderich, L. Lange, H. Adel, J. Strötgen, D. Klakow,
A survey on recent approaches for natural language processing in low-resource scenarios,
\textit{Proceedings of the 2021 Conference of the North American Chapter of the Association for Computational Linguistics: Human Language Technologies},
pp. 2545--2568, 2021.

\bibitem{eisenstein2013bad}
J. Eisenstein,
What to do about bad language on the internet,
\textit{Proceedings of the 2013 Conference of the North American Chapter of the Association for Computational Linguistics: Human Language Technologies},
pp. 359--369, 2013.

\bibitem{waseem2016hateful}
Z. Waseem, D. Hovy,
Hateful symbols or hateful people? Predictive features for hate speech detection on Twitter,
\textit{Proceedings of the NAACL Student Research Workshop},
pp. 88--93, 2016.

\bibitem{aroyo2015truth}
L. Aroyo, C. Welty,
Truth is a lie: Crowd truth and the seven myths of human annotation,
\textit{AI Magazine},
36(1), pp. 15--24, 2015.

\bibitem{kiros2015skip}
R. Kiros, Y. Zhu, R.R. Salakhutdinov, R. Zemel, R. Urtasun, A. Torralba, S. Fidler,
Skip-thought vectors,
\textit{Advances in Neural Information Processing Systems},
pp. 3294--3302, 2015.

\bibitem{joshi2017automatic}
A. Joshi, P. Bhattacharyya, M.J. Carman,
Automatic sarcasm detection: A survey,
\textit{ACM Computing Surveys},
50(5), pp. 1--22, 2017.

\bibitem{sanh2019distilbert}
V. Sanh, L. Debut, J. Chaumond, T. Wolf,
DistilBERT, a distilled version of BERT: smaller, faster, cheaper and lighter,
arXiv preprint arXiv:1910.01108, 2019.

\bibitem{chen2019billion}
T. Chen, M. Li, Y. Li, M. Lin, N. Wang, M. Wang, T. Xiao, B. Xu, C. Zhang, Z. Zhang,
MXNet: A flexible and efficient machine learning library for heterogeneous distributed systems,
arXiv preprint arXiv:1512.01274, 2015.

\bibitem{nissim2020we}
M. Nissim, R. van Noord, R. van der Goot,
Fair is better than sensational: Man is to doctor as woman is to doctor,
\textit{Computational Linguistics},
46(2), pp. 487--497, 2020.

\bibitem{guo2020multi}
J. Guo, D. He, T. Zhao,
Multi-domain neural machine translation,
\textit{IEEE Transactions on Pattern Analysis and Machine Intelligence},
42(7), pp. 1711--1725, 2020.

\bibitem{wang2021multimodal}
Y. Wang, Y. Shen, Z. Liu, P.P. Liang, A. Zadeh, L.P. Morency,
Words can shift: Dynamically adjusting word representations using nonverbal behaviors,
\textit{Proceedings of the AAAI Conference on Artificial Intelligence},
33, pp. 7216--7223, 2019.

\bibitem{watanabe2018espnet}
S. Watanabe, T. Hori, S. Karita, T. Hayashi, J. Nishitoba, Y. Unno, N.E.Y. Soplin, J. Heymann, M. Wiesner, N. Chen, A. Renduchintala, T. Ochiai,
ESPnet: End-to-end speech processing toolkit,
\textit{Proceedings of Interspeech},
pp. 2207--2211, 2018.

\bibitem{poria2017review}
S. Poria, E. Cambria, R. Bajpai, A. Hussain,
A review of affective computing: From unimodal analysis to multimodal fusion,
\textit{Information Fusion},
37, pp. 98--125, 2017.

\bibitem{schuller2013computational}
B. Schuller, S. Steidl, A. Batliner, A. Vinciarelli, K. Scherer, F. Ringeval, M. Chetouani, F. Weninger, F. Eyben, E. Marchi, M. Mortillaro, H. Salamin, A. Polychroniou, F. Valente, S. Kim,
The INTERSPEECH 2013 computational paralinguistics challenge: Social signals, conflict, emotion, autism,
\textit{Proceedings of Interspeech},
2013.

\bibitem{martin2020camembert}
L. Martin, B. Muller, P.J. Ortiz Suárez, Y. Dupont, L. Romary, É. Villemonte de la Clergerie, D. Seddah, B. Sagot,
CamemBERT: a tasty French language model,
\textit{Proceedings of the 58th Annual Meeting of the Association for Computational Linguistics},
pp. 7203--7219, 2020.

\bibitem{gururangan2020dont}
S. Gururangan, A. Marasović, S. Swayamdipta, K. Lo, I. Beltagy, D. Downey, N.A. Smith,
Don't stop pretraining: Adapt language models to domains and tasks,
\textit{Proceedings of the 58th Annual Meeting of the Association for Computational Linguistics},
pp. 8342--8360, 2020.

\bibitem{brown2020language}
T.B. Brown, B. Mann, N. Ryder, M. Subbiah, J. Kaplan, P. Dhariwal, A. Neelakantan, P. Shyam, G. Sastry, A. Askell, S. Agarwal, A. Herbert-Voss, G. Krueger, T. Henighan, R. Child, A. Ramesh, D.M. Ziegler, J. Wu, C. Winter, C. Hesse, M. Chen, E. Sigler, M. Litwin, S. Gray, B. Chess, J. Clark, C. Berner, S. McCandlish, A. Radford, I. Sutskever, D. Amodei,
Language models are few-shot learners,
\textit{Advances in Neural Information Processing Systems},
33, pp. 1877--1901, 2020.

\bibitem{gao2021making}
T. Gao, A. Fisch, D. Chen,
Making pre-trained language models better few-shot learners,
\textit{Proceedings of the 59th Annual Meeting of the Association for Computational Linguistics and the 11th International Joint Conference on Natural Language Processing (Volume 1: Long Papers)},
pp. 3816--3830, 2021.

\bibitem{ju2018exploring}
M. Ju, H. Nguyen, M. Nilsson, C. Tan,
Exploring ensemble methods for multilingual toxic comment classification,
\textit{Proceedings of the First Workshop on Trolling, Aggression and Cyberbullying (TRAC-2018)},
pp. 183--190, 2018.

\bibitem{ruder2019survey}
S. Ruder, I. Vulić, A. Søgaard,
A survey of cross-lingual word embedding models,
\textit{Journal of Artificial Intelligence Research},
65, pp. 569--631, 2019.

\bibitem{lauscher2020common}
A. Lauscher, V. Ravishankar, I. Vulić, G. Glavaš,
From zero to hero: On the limitations of zero-shot language transfer with multilingual transformers,
\textit{Proceedings of the 2020 Conference on Empirical Methods in Natural Language Processing (EMNLP)},
pp. 4483--4499, 2020.

\bibitem{ponti2019modeling}
E.M. Ponti, G. Glavaš, O. Majewska, Q. Liu, I. Vulić, A. Korhonen,
Modeling language variation and universals: A survey on typological linguistics for natural language processing,
\textit{Computational Linguistics},
45(3), pp. 559--601, 2019.

\bibitem{settles2009active}
B. Settles,
Active learning literature survey,
Computer Sciences Technical Report 1648, University of Wisconsin--Madison, 2009.

\bibitem{ein2020active}
D. Ein-Dor, A. Halfon, A. Gera, E. Shnarch, L. Dankin, L. Choshen, M. Danilevsky, R. Aharonov, Y. Katz, N. Slonim,
Active learning for BERT: An empirical study,
\textit{Proceedings of the 2020 Conference on Empirical Methods in Natural Language Processing (EMNLP)},
pp. 7949--7962, 2020.

\bibitem{chen2020lifelong}
Z. Chen, B. Liu,
Lifelong machine learning,
\textit{Synthesis Lectures on Artificial Intelligence and Machine Learning},
12(3), pp. 1--207, 2020.

\bibitem{kirkpatrick2017overcoming}
J. Kirkpatrick, R. Pascanu, N. Rabinowitz, J. Veness, G. Desjardins, A.A. Rusu, K. Milan, J. Quan, T. Ramalho, A. Grabska-Barwinska, D. Hassabis, C. Clopath, D. Kumaran, R. Hadsell,
Overcoming catastrophic forgetting in neural networks,
\textit{Proceedings of the National Academy of Sciences},
114(13), pp. 3521--3526, 2017.

\bibitem{rebuffi2017icarl}
S.A. Rebuffi, A. Kolesnikov, G. Sperl, C.H. Lampert,
iCaRL: Incremental classifier and representation learning,
\textit{Proceedings of the IEEE Conference on Computer Vision and Pattern Recognition},
pp. 2001--2010, 2017.

\end{thebibliography}
\end{document}